\documentclass[11pt]{article}
\usepackage[preprint]{acl}
\usepackage{times}
\usepackage{latexsym}
\usepackage[T1]{fontenc}
\usepackage[utf8]{inputenc}
\usepackage{microtype}
\usepackage{inconsolata}
\usepackage{graphicx}
\usepackage{svg}
\usepackage{amsmath,amssymb,bm}
\usepackage{booktabs,multirow,tabularx,array}

\usepackage{algorithm,algpseudocode}
\usepackage{placeins}
\usepackage{xcolor,tikz}
\usetikzlibrary{positioning,arrows.meta,calc}
\newcommand{\Q}{\mathcal{Q}}
\newcommand{\DQ}{\mathcal{D}}
\newcommand{\R}{\mathbb{R}}

\usepackage[table]{xcolor}
\definecolor{pactblue}{RGB}{230,242,255}

\title{\raisebox{-0.2em}[0pt][0pt]{\includegraphics[height=1.5em]{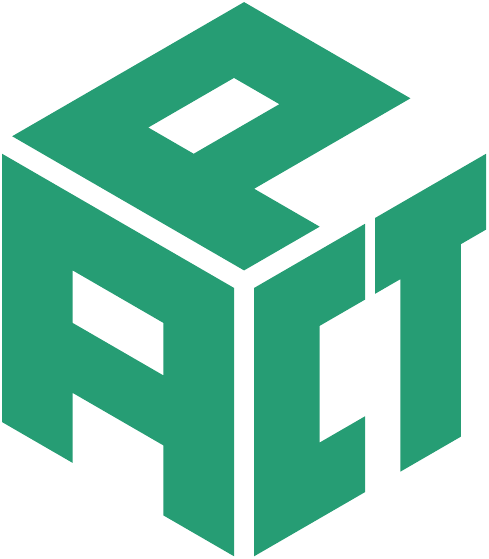}}\hspace{0.35em}PACT: NVFP4 Data-Flow LoRA Fine-Tuning on Consumer GPUs}
\author{%
  Lei Gao, Hossein Entezari Zarch, Chaoyi Jiang, Murali Annavaram\\
  University of Southern California\\
  \texttt{\{leig, entezari, chaoyij, annavara\}@usc.edu}\\
}

\begin{document}
\maketitle

\begin{abstract}
Fine-tuning language models on consumer GPUs is constrained by computation and memory. Although LoRA reduces trainable parameters and optimizer state, frozen projections remain expensive, and activations retained for backward can dominate memory usage. 
We present PACT, a LoRA fine-tuning system that accelerates frozen projections with native NVFP4 GEMMs while reducing quantization and dequantization overhead and limiting quantization error. To reduce activation memory, PACT reuses quantized RMSNorm outputs and saves gate/up activations in NVFP4 for backward. To limit numerical error, it preserves BF16 inputs for forward nonlinear operations and calibrates frozen weights offline using GPTQ adapted to NVFP4 scales and quantized activations. Fused nonlinear kernels and cached forward and backward weight layouts reduce quantization and dequantization overhead.
Evaluations across multiple models and tasks on a 16\,GB RTX~5080 show lower memory usage and faster LoRA fine-tuning with competitive accuracy, enabling model adaptation on resource-constrained consumer hardware.
\end{abstract}

\section{Introduction}
\label{sec:intro}

Large language models (LLMs) are increasingly fine-tuned to adapt to a user's private data, personal writing style, or specialized downstream task~\citep{hu2022lora,gao2025mobizo,tan2024oppu}. Performing this adaptation on the device that holds the data keeps sensitive data local, avoids the latency and recurring cost of cloud round trips, and enables continual learning~\citep{xu2024ondevice,zhu2023pockengine}. In this setting, the practical target is often a small language model (SLM) in the $1$B to $3$B range, which can fit on a consumer GPU before fine-tuning but still faces severe training-time compute and memory pressure.

\begin{figure}[t]
    \centering
    \includegraphics[width=1\linewidth]{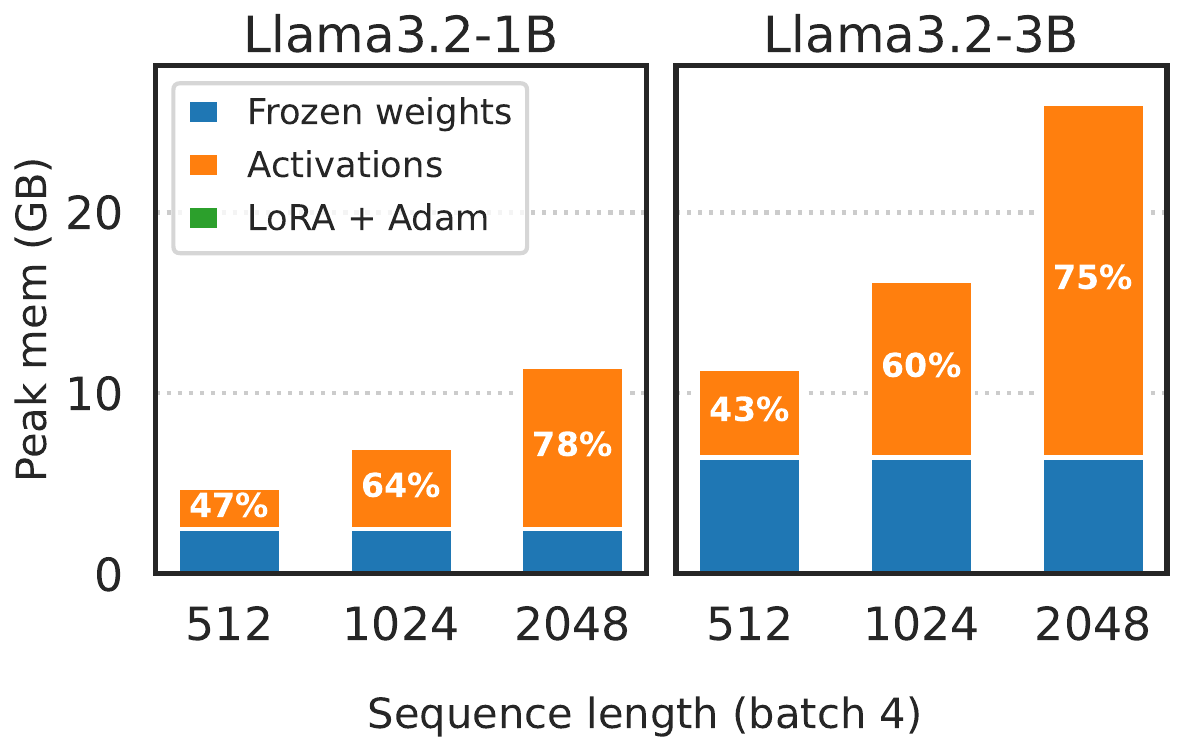}
    \caption{GPU memory composition during BF16 LoRA fine-tuning at batch size 4. Activations dominate memory at longer sequence lengths.}
\label{fig:memory_breakdown}
\end{figure}

Low-rank adaptation (LoRA)~\citep{hu2022lora} reduces the number of trainable parameters by freezing pretrained weights and optimizing small adapter matrices. This substantially reduces gradient and optimizer storage, but the frozen projections are still evaluated in the forward pass and used to propagate input gradients during backward. Standard BF16 LoRA therefore retains substantial computation, while saved activations become a major memory cost as the batch size and sequence length grow. Figure~\ref{fig:memory_breakdown} illustrates this activation-dominated regime on a consumer GPU.

Existing methods address these costs through different mechanisms. QLoRA~\citep{dettmers2023qlora} compresses frozen weights to four bits while retaining higher-precision computation and activations. Mixed-precision training accelerates matrix multiplications by quantizing their operands~\citep{deepseekai2024deepseekv3,nvidia2025pretrainingnvfp4}, but it does retain quantized activations for backward computation. Consequently, retained activations can remain a substantial memory cost even when weights and computation use low precision. These approaches motivate jointly designing the computation and storage of a fine-tuning system.

NVIDIA Blackwell GPUs provide native NVFP4 matrix multiplication, offering a hardware basis for this design. At batch size 4 and sequence length 1,024, our representative linear-layer measurements show that NVFP4 GEMMs are approximately $6$--$7\times$ faster than BF16 GEMMs (Table~\ref{tab:fp4_speedup}). Realizing this benefit during training requires carefully designing the data flow to reduce the overhead of quantization and dequantization operations while limiting quantization error across both the forward and backward passes.

\begin{table}[h]
\centering
\resizebox{\columnwidth}{!}{%
\begin{tabular}{lcccc}
\toprule
\textbf{Layer} & $d_{\text{in}}\times d_{\text{out}}$ & \textbf{BF16} & \textbf{NVFP4} & \textbf{Speedup} \\
\midrule
query   & $3072 \times 3072$  & 0.690 ms & 0.113 ms & $6.1\times$ \\
gate/up   & $3072 \times 8192$  & 1.778 ms & 0.304 ms & $5.8\times$ \\
down      & $8192 \times 3072$  & 1.818 ms & 0.267 ms & $6.8\times$ \\
\bottomrule
\end{tabular}
}
\caption{GEMM latency for representative Llama-3B projections. NVFP4 provides approximately $6$--$7\times$ acceleration over BF16 on a RTX 5080 GPU.}
\label{tab:fp4_speedup}
\end{table}

We introduce \textbf{PACT}, an NVFP4-based LoRA fine-tuning system that combines accelerated frozen projections with compressed activation storage. To accelerate computation, PACT uses native NVFP4 GEMMs for the forward and input-gradient computations of all seven frozen projections in each decoder layer. To reduce activation memory, PACT reuses quantized RMSNorm outputs for backward and saves separate NVFP4 copies of the gate/up activations. To reduce quantization error, SiLU-multiply uses the original BF16 gate/up values in forward, which are released after use. To reduce overhead caused by quantization and dequantization operations, fused kernels are implemented to produce both the compressed backward state and the quantized inputs required by subsequent projections, avoiding additional GPU memory traffic.

Frozen base weights provide a complementary opportunity to further reduce quantization error at the offline phase. PACT adapts GPTQ~\citep{frantar2023gptq} to NVFP4 by evaluating representable block scales and constructing calibration statistics from quantized activations. This aligns weight calibration with the activation precision used during training. To further improve computation efficiency, PACT caches quantized weight representations in the layouts required by forward and input-gradient GEMMs, avoiding repeated weight quantization during training.

Our contributions are:
\begin{itemize}
    \item We design an NVFP4 LoRA data flow that combines accelerated frozen projections with compressed activation storage. It reuses quantized RMSNorm outputs and saves gate/up activations in NVFP4 for backward while preserving BF16 inputs for forward nonlinear computation.
    \item We reduce quantization and dequantization overhead through fused Triton kernels and cached weight layouts. We also adapt GPTQ to NVFP4 block scales and quantized calibration inputs to reduce weight quantization error offline.
    \item We evaluate PACT across language and vision-language tasks on a 16\,GB consumer GPU. Compared with BF16 LoRA, PACT reduces peak memory by 47\% to 51\% and achieves $1.7\times$ to $1.9\times$ training-step speedups at matched batch shapes, with an average classification accuracy gap of at most 2.6 percentage points.
\end{itemize}

\section{Background}
\label{sec:background}

\textbf{Quantization.}
Quantization~\citep{gholami2021survey,nagel2021whitepaper} reduces numerical precision by mapping a high-precision tensor $\mathbf{X}$ to a finite set of discrete levels, thereby lowering memory footprint and computational cost. Formally, for a target integer or floating-point type with $k$-bit codes having dynamic range $[-Q_{\max}, Q_{\max}]$, each element is quantized as
\begin{equation}
    \tilde{x}_i = \operatorname{round}_{\mathcal{C}}\!\left(\frac{x_i}{s}\right), \qquad s = \frac{\max(|\mathbf{X}|)}{Q_{\max}},
    \label{eq:quantization}
\end{equation}
where $s$ is the per-tensor scaling factor and $\operatorname{round}_{\mathcal{C}}$ rounds to the nearest representable value in the target codebook $\mathcal{C}$. The code for $\tilde{x}_i$ is stored in $k$ bits.

\textbf{Block-scaled quantization.}
The NVIDIA Blackwell architecture~\citep{nvidiaBlackwellArchitecture} introduces hardware acceleration for block-scaled formats. Unlike per-tensor methods, block-scaled quantization operates on small groups of elements, each with its own scaling factor. Specifically, MXFP4 groups 32 E2M1 elements under one E8M0 scale~\citep{rouhani2023microscaling}, whereas NVFP4 groups 16 E2M1 elements under one E4M3 scale~\citep{nvidia2025pretrainingnvfp4}. Although Blackwell supports both formats, this work focuses on NVFP4 for its finer-grained block size and higher-precision scaling factor. A detailed comparison is provided in Appendix~\ref{app:fp4_compare}.

\textbf{Low-rank adaptation (LoRA).}
LoRA~\citep{hu2022lora} is a parameter-efficient fine-tuning method that freezes the pretrained weight $\mathbf{W}_0 \in \R^{d_{\text{out}} \times d_{\text{in}}}$ and models its update $\Delta \mathbf{W} \in \R^{d_{\text{out}} \times d_{\text{in}}}$ as a low-rank product,
$
    \mathbf{W}' = \mathbf{W}_0 + \Delta \mathbf{W} = \mathbf{W}_0 + \mathbf{B}\mathbf{A},
$
where $\mathbf{A} \in \R^{r \times d_{\text{in}}}$ and $\mathbf{B} \in \R^{d_{\text{out}} \times r}$ are the trainable low-rank factors with $r \ll \min(d_{\text{in}}, d_{\text{out}})$, while $\mathbf{W}_0$ remains frozen throughout training. The forward pass is computed as
\begin{equation}
    \mathbf{Y} = \mathbf{X}\mathbf{W}_0^\top + \mathbf{X}\mathbf{A}^\top\mathbf{B}^\top,
    \label{eq:lora_forward}
\end{equation}
where $\mathbf{A}$ is initialized randomly and $\mathbf{B}$ is initialized to zero, so that $\Delta \mathbf{W} = \mathbf{0}$ at the start of fine-tuning and the pretrained behavior is preserved.

\section{PACT}
\label{sec:method}

\begin{figure}[t!]
    \centering
    \includegraphics[width=\columnwidth]{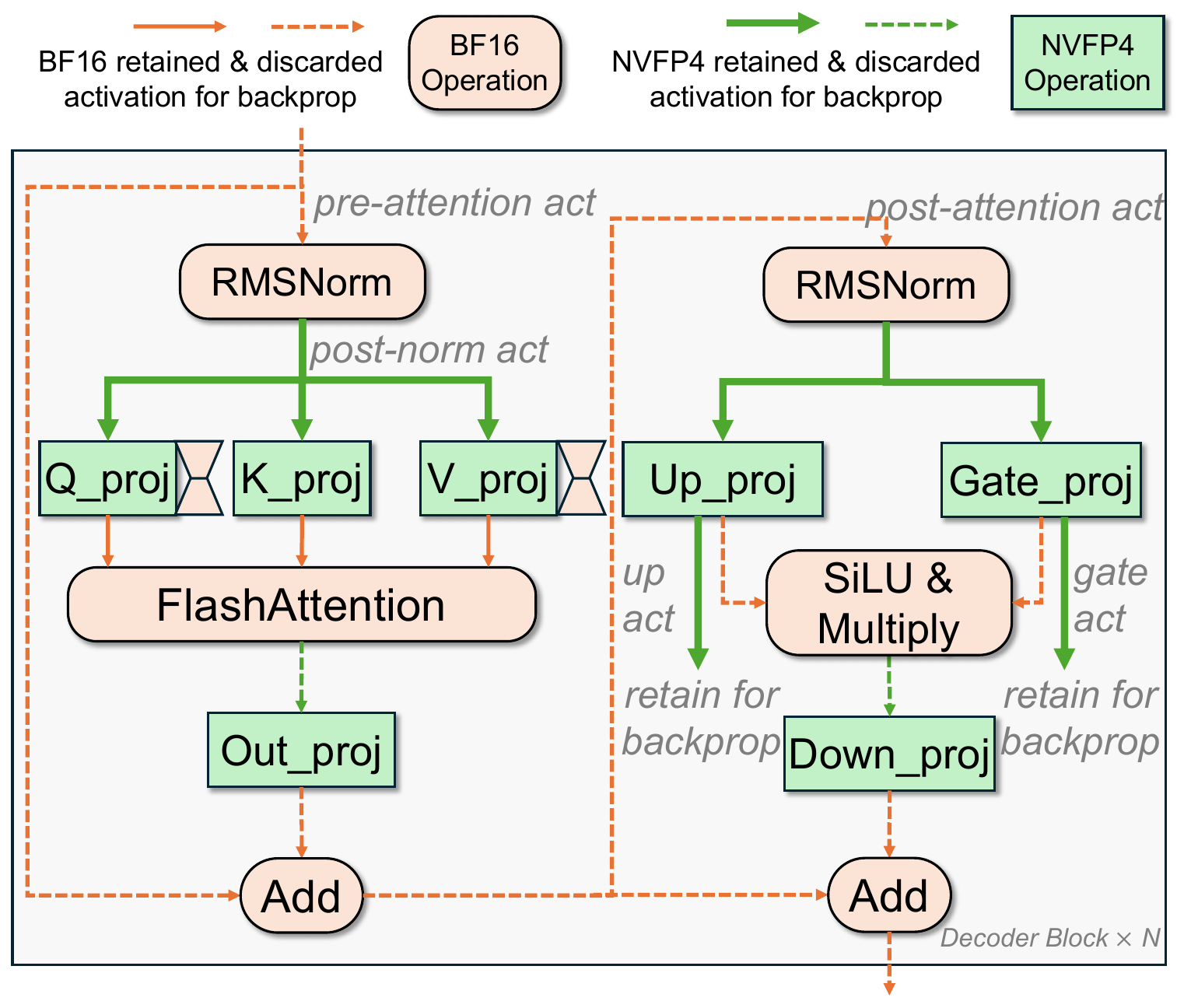}
    \caption{PACT decoder data flow. Native NVFP4 projections accelerate computation, while NVFP4 storage reduces retained activations. Attention backward state remains in BF16.}
\label{fig:method}
\end{figure}

PACT uses native NVFP4 GEMMs to accelerate frozen projections (Section~\ref{sec:gemm}) and stores selected activations in NVFP4 for backward (Section~\ref{sec:state}), as shown in Figure~\ref{fig:method}. To limit quantization error, it preserves BF16 inputs for forward SiLU-multiply and calibrates frozen weights offline (Section~\ref{sec:weights}). Fused kernels combine nonlinear computation with quantization and dequantization to reduce conversion overhead and intermediate memory traffic. Cached forward and backward weight layouts further avoid repeated weight quantization during training.

\subsection{NVFP4 GEMM Acceleration}
\label{sec:gemm}

\textbf{Linear layers.}
To accelerate both forward and backward computation, PACT executes all seven frozen projections in each decoder layer with native NVFP4 GEMMs, including query, key, value, attention output, gate, up, and down. For a frozen weight $\mathbf{W}_0$ and input $\mathbf{X}$, the forward and input-gradient computations are
\begin{align}
    \underbrace{
        \mathbf{Y}\vphantom{\tilde{\mathbf{X}}\tilde{\mathbf{W}}_0^\top}
    }_{\text{BF16}}
    &= \underbrace{
        \tilde{\mathbf{X}}\tilde{\mathbf{W}}_0^\top
    }_{\text{NVFP4}\,\times\,\text{NVFP4}},\\
    \underbrace{
        \nabla\mathbf{X}
        \vphantom{\Q(\nabla\mathbf{Y})\tilde{\mathbf{W}}_{0,b}}
    }_{\text{BF16}}
    &\simeq \underbrace{
        \Q(\nabla\mathbf{Y})\tilde{\mathbf{W}}_{0,b}
    }_{\text{NVFP4}\,\times\,\text{NVFP4}}.
    \label{eq:base_gemms}
\end{align}
We use $\tilde{\mathbf{X}}=\Q(\mathbf{X})$ for a quantized tensor and $\bar{\mathbf{X}}=\DQ(\tilde{\mathbf{X}})$ for its reconstruction. The backward weight $\tilde{\mathbf{W}}_{0,b}$ is prepared separately as described in Section~\ref{sec:weights}. Quantized operands include their scales. Both GEMMs accumulate in FP32 and return BF16 outputs.

To reduce quantization overhead, PACT fuses quantization into the RMSNorm and SiLU-multiply kernels that produce the projection inputs. These kernels read BF16 inputs, compute in FP32, and write NVFP4 outputs directly. This avoids a separate quantization kernel and the additional write and read of a higher-precision output. Before each input-gradient GEMM, PACT quantizes the BF16 output gradient using an FP32 tensor scale and unsigned E4M3 (UE4M3) block scales. Appendix~\ref{app:quantizer} gives the scaling details.

\begin{figure}[t!]
    \centering
    \includegraphics[width=\columnwidth]{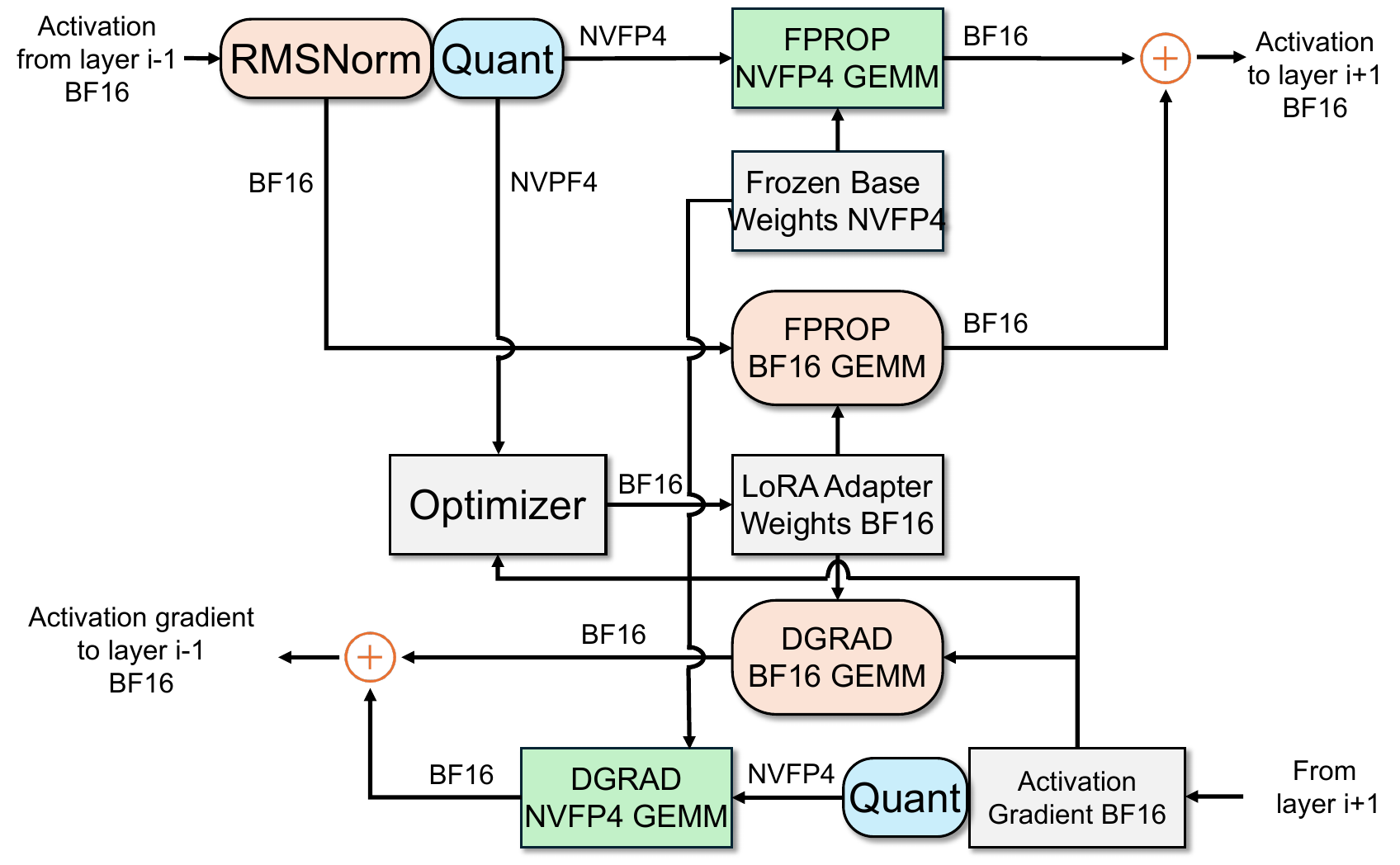}
    \caption{PACT LoRA linear-layer forward and backward paths, using NVFP4 for frozen base weights and BF16 for trainable adapter computation.}
\label{fig:lora}
\end{figure}

\textbf{LoRA layers.}
As illustrated in Figure~\ref{fig:lora}, the LoRA branch reuses the reconstructed input $\bar{\mathbf{X}}$ and keeps its trainable factors $\mathbf{A},\mathbf{B}$ in BF16. With the adapter scaling factor set to one, as in our experiments, the forward pass is
\begin{equation}
    \mathbf{Y}
    = \underbrace{\tilde{\mathbf{X}}\tilde{\mathbf{W}}_0^\top}_{\text{NVFP4 GEMM}}
    + \underbrace{(\bar{\mathbf{X}}\mathbf{A}^\top)\mathbf{B}^\top}_{\text{BF16 GEMMs}}.
    \label{eq:pact_lora_forward}
\end{equation}
Backward computes
\begin{align}
    \nabla\mathbf{X}
    &\simeq \Q(\nabla\mathbf{Y})\tilde{\mathbf{W}}_{0,b}
       +(\nabla\mathbf{Y}\mathbf{B})\mathbf{A},\label{eq:pact_dx}\\
    \nabla\mathbf{B}
    &=(\nabla\mathbf{Y})^\top(\bar{\mathbf{X}}\mathbf{A}^\top),\label{eq:pact_dB}\\
    \nabla\mathbf{A}
    &=(\nabla\mathbf{Y}\mathbf{B})^\top\bar{\mathbf{X}}.\label{eq:pact_dA}
\end{align}
The base input-gradient term uses NVFP4, while the adapter terms use BF16. The BF16 GEMMs are evaluated in the order indicated by the parentheses, avoiding the need to form the full weight update matrix $\mathbf{B}\mathbf{A}$. Attention and residual operations remain in BF16, with the complete precision allocation given in Appendix~\ref{app:precision} and the operator algorithms in Appendix~\ref{app:algorithms}.

\subsection{NVFP4 Activation Storage Reduction}
\label{sec:state}

Beyond accelerating GEMMs, PACT reduces memory by storing selected backward state in NVFP4. RMSNorm reuses its quantized output, while SiLU-multiply uses BF16 gate/up inputs in forward and retains separate NVFP4 copies for backward. Customized RMSNorm and SiLU-multiply kernels limit the overhead of storing and reconstructing these compressed activations.

\textbf{Reuse RMSNorm outputs for backward.}
At the pre-attention RMSNorm, standard LoRA training saves the normalization input for RMSNorm backward and its output for adapter weight gradients. PACT instead reuses the NVFP4 output already produced for the Q/K/V projections. Together with a per-token normalization statistic, this output supports both adapter weight gradients and approximate RMSNorm backward, eliminating the separately saved input. The pre-MLP RMSNorm likewise retains its NVFP4 output for normalization backward.

Specifically, let $\mathbf{y}=\boldsymbol{\gamma}\odot\hat{\mathbf{x}}$, where $\hat{\mathbf{x}}=r_x\mathbf{x}$ and $r_x$ is the inverse root-mean-square statistic. For elementwise nonzero $\boldsymbol{\gamma}$, the normalized input can be approximated as $\hat{\mathbf{x}}\simeq\bar{\mathbf{y}}/\boldsymbol{\gamma}$. Using the reconstructed output $\bar{\mathbf{y}}$ and the saved FP32 statistic $r_x$, backward computes
\begin{equation}
\nabla\mathbf{x}\simeq r_x\left[\boldsymbol{\gamma}\odot\nabla\mathbf{y}-
\frac{\bar{\mathbf{y}}}{\boldsymbol{\gamma}}\operatorname{mean}(\bar{\mathbf{y}}\odot\nabla\mathbf{y})\right].
\label{eq:rms_backward}
\end{equation}

\textbf{Compress MLP intermediates for backward.}
SiLU-multiply backward requires the gate and up activations. To compress this state without adding quantization error to the forward inputs, PACT evaluates SiLU-multiply using the original BF16 values and saves separate NVFP4 copies. A fused kernel reads the BF16 gate/up values $\mathbf{g},\mathbf{u}$, computes $\mathbf{z}=\operatorname{SiLU}(\mathbf{g})\odot\mathbf{u}$ in FP32, and quantizes $\mathbf{z}$ for the down projection. It also writes the NVFP4 gate/up copies in the same pass, avoiding separate kernels for saved-state quantization. The BF16 originals can then be released.

During backward, a fused kernel reconstructs the saved NVFP4 values on chip and computes approximate gradients
\begin{align}
\nabla\mathbf{g}&\simeq\nabla\mathbf{z}\odot\bar{\mathbf{u}}\odot\operatorname{SiLU}'(\bar{\mathbf{g}}),\\
\nabla\mathbf{u}&\simeq\nabla\mathbf{z}\odot\operatorname{SiLU}(\bar{\mathbf{g}}).
\end{align}
Fusing dequantization with derivative computation reduces conversion overhead by avoiding separate writes and reads of reconstructed gate/up tensors in GPU memory. The kernel returns BF16 gradients, which are subsequently quantized for the gate and up input-gradient GEMMs.

\textbf{Analytical activation storage reduction.}
Within each decoder layer, PACT stores the pre-attention and pre-MLP RMSNorm outputs and the gate/up activations in NVFP4, while retaining the attention state in BF16. These choices yield the following estimate of activation storage relative to BF16 LoRA.
Let $U=2bsd$ bytes denote the storage required for one BF16 hidden-state tensor, where $b$, $s$, and $d$ are the batch size, sequence length, and hidden dimension, respectively. Including block scales, NVFP4 requires approximately 28\% of the storage of BF16. Using Llama-3B dimensions and counting the attention query, key, value, and output tensors at hidden width, the retained activation storage per layer is approximately
\begin{equation}
\underbrace{0.56U}_{\text{two norm outputs}}+
\underbrace{1.50U}_{\text{gate/up}}+
\underbrace{4U}_{\text{attention}}
\simeq6.1U.
\label{eq:memory}
\end{equation}
Under the same accounting, BF16 LoRA fine-tuning requires approximately $12.3U$, yielding an estimated $2\times$ reduction in retained activation memory.

\subsection{NVFP4 Weight Quantization and Calibration}
\label{sec:weights}

Frozen weights allow PACT to further reduce quantization error before training. PACT adapts GPTQ by selecting representable NVFP4 block scales and constructing calibration statistics from quantized activations. It then caches the calibrated weights in the layouts required by forward and input-gradient GEMMs. Since the weights remain fixed, both calibration and layout preparation are performed only once.

\textbf{NVFP4-aware scale selection.}
NVFP4 represents weights using the E2M1 codebook, an FP32 tensor scale $s_W$, and UE4M3 scales shared by contiguous groups of 16 weights. Scale selection must therefore account for rounding in both the weight values and their block scales. Larger scales increase the representable range, while smaller scales improve resolution at the cost of clipping.

PACT evaluates candidate scales after rounding them to UE4M3, measuring reconstruction error with E2M1 weights. For $\mathbf{W}_0\in\R^{d_{\text{out}}\times d_{\text{in}}}$, groups follow the forward reduction dimension $d_{\text{in}}$. Given a current, error-compensated group $\mathbf{w}_j$, a candidate factor $f$ defines
\begin{align}
s_j(f)&=\operatorname{UE4M3}\!\left(f\frac{\max|\mathbf{w}_j|}{6s_W}\right),\\
\hat{\mathbf{w}}_j(f)&=s_Ws_j(f)\,\operatorname{E2M1}\!\left(\frac{\mathbf{w}_j}{s_Ws_j(f)}\right).
\end{align}
Here, $\operatorname{UE4M3}$ and $\operatorname{E2M1}$ round to the corresponding representable values. We choose the candidate that minimizes $\|\mathbf{w}_j-\hat{\mathbf{w}}_j(f)\|_2^2$ and hold its scale fixed during subsequent column quantization. The candidate set covers both scale expansion and clipping. Appendix~\ref{app:quantizer} provides the search factors and tensor-scale definition.

\textbf{Calibration under quantized activations.}
To account for activation quantization during weight calibration, PACT constructs GPTQ's calibration statistics from reconstructed NVFP4 inputs $\bar{\mathbf{X}}=\DQ(\Q(\mathbf{X}))$. This aligns the calibration objective with the activation precision used during training:
\begin{align}
\min_{\hat{\mathbf{W}}_0\in\mathcal{W}_{\mathrm{NVFP4}}}
&\,\|\bar{\mathbf{X}}\mathbf{W}_0^\top-\bar{\mathbf{X}}\hat{\mathbf{W}}_0^\top\|_F^2,\\
\mathbf{H}&=\frac{2}{m}\bar{\mathbf{X}}^\top\bar{\mathbf{X}},
\label{eq:hessian}
\end{align}
where $m$ is the number of calibration token rows and $\mathcal{W}_{\mathrm{NVFP4}}$ denotes the set of representable weight tensors. To reduce this output discrepancy, PACT quantizes columns sequentially and compensates the remaining weights for each column's quantization error using inverse-Hessian updates. Both scale selection and error compensation therefore account for the numerical representation used by PACT's forward GEMMs.

\textbf{Forward and backward weight layouts.}
The forward GEMM uses NVFP4 $\tilde{\mathbf{W}}_0^\top$, while the input-gradient GEMM uses a separately quantized representation $\tilde{\mathbf{W}}_{0,b}$. However, NVFP4 groups weights along the GEMM reduction dimension, which changes from $d_{\text{in}}$ in forward to $d_{\text{out}}$ in backward. Simply transposing the packed forward representation therefore does not produce the required backward layout. PACT reconstructs and transposes the calibrated forward weights, then requantizes them into NVFP4 along the new reduction dimension using separate block and tensor scales. Caching both layouts removes weight transposition and requantization from the training loop, reducing conversion overhead in both forward and backward computation.

\section{Experiments}
\label{sec:experiments}

We evaluate PACT's adaptation quality, training speed, and memory usage, and compare it with alternative approaches to reducing activation memory.

\begin{table*}[!t]
\centering\small
\setlength{\tabcolsep}{3.3pt}
\begin{tabular}{llrrrrrrrrrr}
\toprule
Model & Method & SST-2 & RTE & BoolQ & WSC & WiC & COPA & WinoG. & ARC-E & ARC-C & Avg. \\
\midrule
\multirow{5}{*}{Llama-1B}
& Zero-shot & .589 & .520 & .584 & .365 & .500 & .730 & .575 & .472 & .313 & .516 \\
& BF16 LoRA & .946 & .816 & .811 & .635 & .705 & .777 & .645 & .755 & .475 & .729 \\
& QLoRA     & .944 & .786 & .817 & .635 & .705 & .767 & .622 & .728 & .433 & .715 \\
& TE        & .932 & .785 & .802 & .571 & .657 & .793 & .602 & .698 & .372 & .690 \\
\rowcolor{pactblue}
& PACT      & .940 & .823 & .790 & .551 & .665 & .777 & .612 & .729 & .444 & .703 \\
\midrule
\multirow{5}{*}{Llama-3B}
& Zero-shot & .589 & .473 & .680 & .413 & .500 & .780 & .661 & .803 & .667 & .618 \\
& BF16 LoRA & .955 & .889 & .876 & .625 & .722 & .870 & .752 & .866 & .738 & .810 \\
& QLoRA     & .955 & .862 & .875 & .641 & .714 & .877 & .736 & .860 & .712 & .804 \\
& TE        & .953 & .866 & .870 & .615 & .708 & .867 & .713 & .844 & .690 & .792 \\
\rowcolor{pactblue}
& PACT      & .950 & .866 & .869 & .638 & .721 & .870 & .737 & .846 & .707 & .801 \\
\midrule
\multirow{5}{*}{Qwen2.5-1.5B}
& Zero-shot & .573 & .805 & .717 & .471 & .558 & .770 & .598 & .869 & .707 & .674 \\
& BF16 LoRA & .945 & .863 & .841 & .619 & .724 & .830 & .660 & .906 & .816 & .800 \\
& QLoRA     & .944 & .842 & .843 & .625 & .700 & .813 & .638 & .904 & .786 & .788 \\
& TE        & .935 & .839 & .825 & .571 & .714 & .797 & .634 & .878 & .781 & .775 \\
\rowcolor{pactblue}
& PACT      & .941 & .853 & .837 & .615 & .703 & .823 & .625 & .899 & .782 & .787 \\
\bottomrule
\end{tabular}
\caption{Classification and multiple-choice accuracy across nine tasks. PACT retains most of the BF16 fine-tuning gains and exceeds TE in average accuracy on all three models.}
\label{tab:accuracy}
\end{table*}

\subsection{Models, Protocol, and Baselines}
\label{sec:setup}

\textbf{Models and datasets.}
We study three open-weight LLMs with $1$ to $3$B parameters: Llama-3.2-1B, Llama-3.2-3B~\citep{grattafiori2024llama3}, and Qwen2.5-1.5B~\citep{yang2024qwen25}. We abbreviate the two Llama models as Llama-1B and Llama-3B. Our classification and multiple-choice benchmarks include SST-2 for sentiment classification~\citep{wang2019glue}, ARC-Easy and ARC-Challenge for science question answering~\citep{clark2018arc}, and WinoGrande for commonsense reasoning~\citep{sakaguchi2021winogrande}. We also evaluate five SuperGLUE tasks~\citep{wang2019superglue}: BoolQ for reading comprehension, RTE for textual entailment, WSC for coreference resolution, WiC for word-sense disambiguation, and COPA for causal reasoning. To evaluate open-ended generation, we use SQuAD for reading comprehension~\citep{rajpurkar2016squad}, GSM8K for mathematical reasoning~\citep{cobbe2021gsm8k}, and SAMSum for dialogue summarization~\citep{gliwa2019samsum}. For vision-language adaptation, we fine-tune the language decoder of Qwen2.5-VL-3B-Instruct~\citep{bai2025qwen25vl} on the image-based questions in ScienceQA~\citep{lu2022scienceqa}.

\textbf{Baselines.}
We compare PACT with four baselines: zero-shot evaluation, BF16 LoRA as the higher-precision reference, QLoRA with $4$-bit NF4 frozen weights, BF16 activations and no gradient checkpointing, and NVFP4 TE. NVFP4 TE is NVIDIA's TransformerEngine~\cite{nvidiaTransformerEngine2026} implementation of NVFP4 mixed-precision pretraining, which we modify to support LoRA fine-tuning. Its linear GEMMs run on NVFP4 tensor cores, while non-linear operators, the LoRA branch, and saved activations remain in BF16.

\textbf{Hyperparameters and hardware.}
For the matched classification experiments, we use rank-$16$ LoRA adapters on the query and value projections and train for up to $500$ optimizer steps or two epochs, whichever comes first. For each task and method, we sweep learning rates $\{10^{-5},5\!\times\!10^{-5},10^{-4},5\!\times\!10^{-4}\}$ and select the rate with the highest mean accuracy across seeds 42, 43, and 44. All main experiments use PyTorch~$2.11.0$, FlashAttention~$2.8.3$, and CUDA~$13.2$ on a single RTX~5080 GPU with $16$\,GB of memory. Appendix~\ref{app:protocol} provides the task-specific training and evaluation settings.

\subsection{Adaptation Quality}
\label{sec:accuracy}

\textbf{Classification.}
Table~\ref{tab:accuracy} shows that PACT preserves substantial fine-tuning gains, improving average accuracy over zero-shot evaluation by more than 11 percentage points across the three models. Its average accuracy exceeds TE on all three models and remains within 0.3 points of QLoRA on Llama-3B and Qwen2.5-1.5B, despite also quantizing activations and input-gradient GEMM operands. Relative to BF16 LoRA, the average gap is 0.9 to 1.3 points on these two models and 2.6 points on Llama-1B. Sensitivity remains task-dependent: Llama-1B shows larger losses on WSC and WiC, while slightly improving on RTE and matching BF16 on COPA.

\begin{table}[!t]
\centering\small
\setlength{\tabcolsep}{2.5pt}
\begin{tabular}{llrrr}
\toprule
Model & Method & SQuAD & GSM8K & SAMSum \\
& & F1 & EM & R-L \\
\midrule
\multirow{4}{*}{Llama-1B}
& BF16 LoRA & .806 & .170 & .418 \\
& QLoRA & .793 & .161 & .413 \\
& TE & .761 & .131 & .405 \\
\rowcolor{pactblue}
& PACT & .782 & .148 & .408 \\
\midrule
\multirow{4}{*}{Llama-3B}
& BF16 LoRA & .837 & .393 & .444 \\
& QLoRA & .845 & .373 & .442 \\
& TE & .832 & .338 & .435 \\
\rowcolor{pactblue}
& PACT & .842 & .371 & .444 \\
\midrule
\multirow{4}{*}{Qwen2.5-1.5B}
& BF16 LoRA & .833 & .606 & .423 \\
& QLoRA & .822 & .565 & .423 \\
& TE & .797 & .493 & .417 \\
\rowcolor{pactblue}
& PACT & .824 & .585 & .416 \\
\bottomrule
\end{tabular}
\caption{Generation quality on SQuAD, GSM8K, and SAMSum. PACT exceeds TE on eight of nine model and task combinations.}
\label{tab:generation}
\end{table}

\begin{figure*}[!t]
\centering
\includegraphics[width=\textwidth]{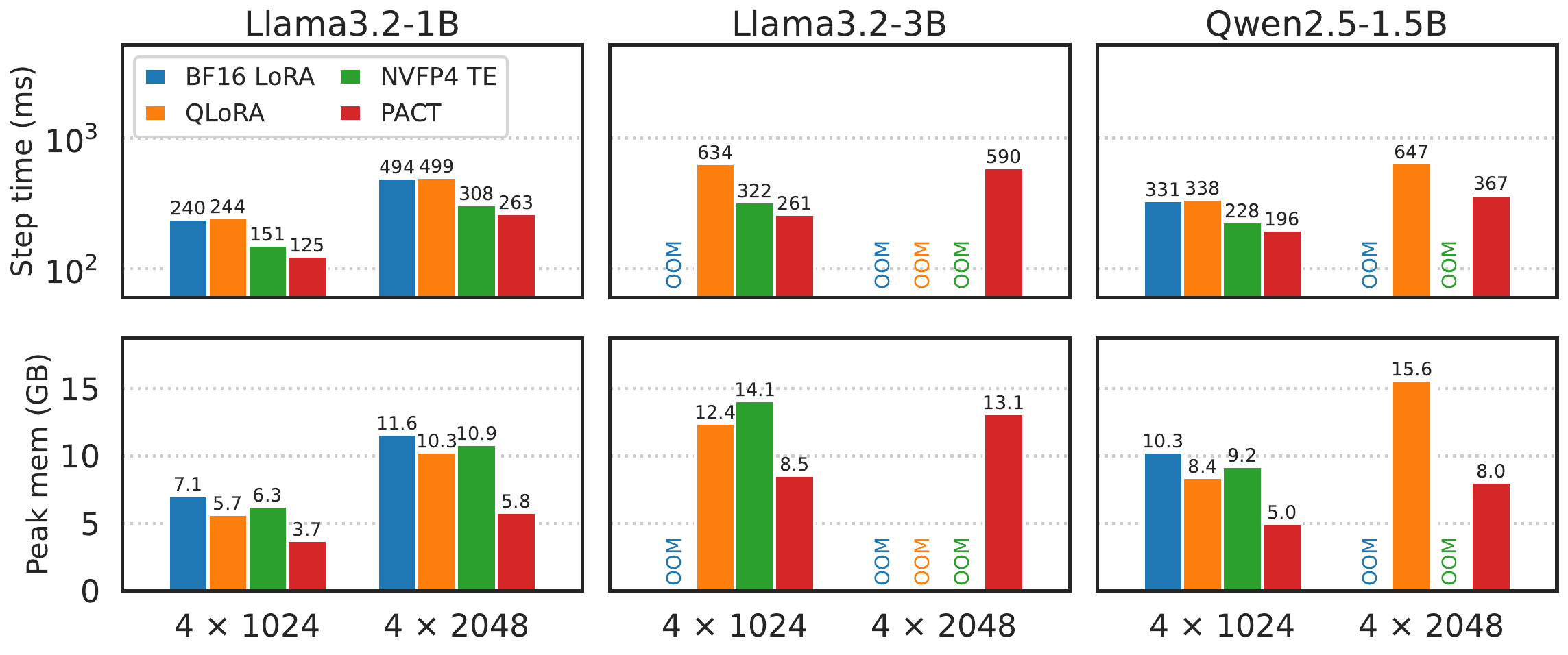}
\caption{Training-step time and peak memory at fixed batch size and sequence length. PACT lowers both costs and fits workloads that exceed the memory budget of other methods.}
\label{fig:fixed}
\end{figure*}

\textbf{Open-ended generation.}
Table~\ref{tab:generation} reports SQuAD F1, GSM8K exact match, and SAMSum ROUGE-L. PACT outperforms TE on eight of nine model and task combinations and generally approaches QLoRA. On Llama-3B, it slightly exceeds BF16 on SQuAD (0.842 vs.\ 0.837) and matches it on SAMSum, although both tasks show modest gaps on the smaller models. GSM8K remains more sensitive, with PACT trailing BF16 by about two percentage points across all three models. Nevertheless, on Qwen2.5-1.5B, PACT exceeds QLoRA and TE on GSM8K by 2.0 and 9.2 points, respectively.

\begin{table}[!tb]
\centering\small
\begin{tabular}{lrr}
\toprule
Qwen2.5-VL-3B / ScienceQA & Before & After \\
\midrule
BF16 LoRA & .779 & .936 \\
PACT, text calibration & .754 & .876 \\
PACT, image-prompt calibration & .762 & .917 \\
\bottomrule
\end{tabular}
\caption{ScienceQA image-question accuracy before and after adaptation. Image-prompt calibration narrows the gap between PACT and BF16 LoRA.}
\label{tab:vlm}
\end{table}

\textbf{Vision-language adaptation.}
We extend PACT to the language decoder of Qwen2.5-VL-3B while keeping the vision encoder and merger frozen in BF16. On ScienceQA image questions (Table~\ref{tab:vlm}), calibrating with 128 training image prompts instead of WikiText improves accuracy by 4.1 percentage points to 0.917, bringing PACT within 1.9 points of BF16 LoRA. These results demonstrate that PACT also supports effective multimodal adaptation when calibrated with representative inputs.

\subsection{Memory and Throughput}
\label{sec:systems}

\textbf{Fixed-shape execution.}
Figure~\ref{fig:fixed} compares training-step time and peak memory at batch size 4.  At sequence length 1,024, PACT achieves $1.7\times$ to $1.9\times$ speedups over BF16 LoRA on Llama-1B and Qwen2.5-1.5B while reducing peak memory by 47\% to 51\%. It also enables Llama-3B training where BF16 runs out of memory. Across all three models, PACT outperforms QLoRA and TE in both speed and memory usage. These gains come from jointly reducing computation and activation storage: QLoRA compresses frozen weights but retains BF16 GEMMs and activations, while TE accelerates GEMMs but retains higher-precision saved state.

The memory savings also enable longer sequences. At length 2,048, PACT fits all three models and is the only evaluated method that fits Llama-3B at batch 4, using 13.11\,GB. On Qwen2.5-1.5B, it uses roughly half the memory of QLoRA (8.04 vs.\ 15.61\,GB). These measurements include weights, activations, and temporary buffers, whereas Equation~\ref{eq:memory} estimates only retained activation storage.

\begin{table}[!t]
\centering\small
\setlength{\tabcolsep}{3.5pt}
\begin{tabular}{llrr}
\toprule
Model & Method & Tested batch & Tokens/s \\
\midrule
\multirow{4}{*}{Llama-1B}
& BF16 LoRA & 10 & 17,162 \\
& QLoRA & 12 & 16,800 \\
& TE & 12 & 27,113 \\
\rowcolor{pactblue}
& PACT & 26 & 31,723 \\
\midrule
\multirow{4}{*}{Llama-3B}
& BF16 LoRA & 2 & 6,217 \\
& QLoRA & 4 & 6,454 \\
& TE & 4 & 12,738 \\
\rowcolor{pactblue}
& PACT & 10 & 14,985 \\
\midrule
\multirow{4}{*}{Qwen2.5-1.5B}
& BF16 LoRA & 4 & 12,365 \\
& QLoRA & 8 & 13,140 \\
& TE & 4 & 17,935 \\
\rowcolor{pactblue}
& PACT & 16 & 22,949 \\
\bottomrule
\end{tabular}
\caption{Throughput at each method's largest tested fitting batch at sequence length 1,024. PACT supports larger batches and achieves the highest throughput on all three models.}
\label{tab:capacity}
\end{table}

\textbf{Capacity and throughput.}
Table~\ref{tab:capacity} shows how PACT's memory savings translate into larger batches and higher throughput at sequence length 1,024. PACT supports batches of 26, 10, and 16 on Llama-1B, Llama-3B, and Qwen2.5-1.5B, respectively, compared with 10, 2, and 4 for BF16 LoRA. At each method's largest tested fitting batch, PACT achieves $1.85\times$ to $2.41\times$ the throughput of BF16, $1.75\times$ to $2.32\times$ that of QLoRA, and $1.17\times$ to $1.28\times$ that of TE. These gains combine faster execution with increased batch capacity. The reported batches are the largest found by a discrete search.
Appendix~\ref{app:memory_alternatives} compares PACT with other activation memory reduction methods such as gradient checkpointing and HyC-LoRA~\citep{hyclora2025}.

\begin{figure}[!t]
\centering
\includegraphics[width=\columnwidth]{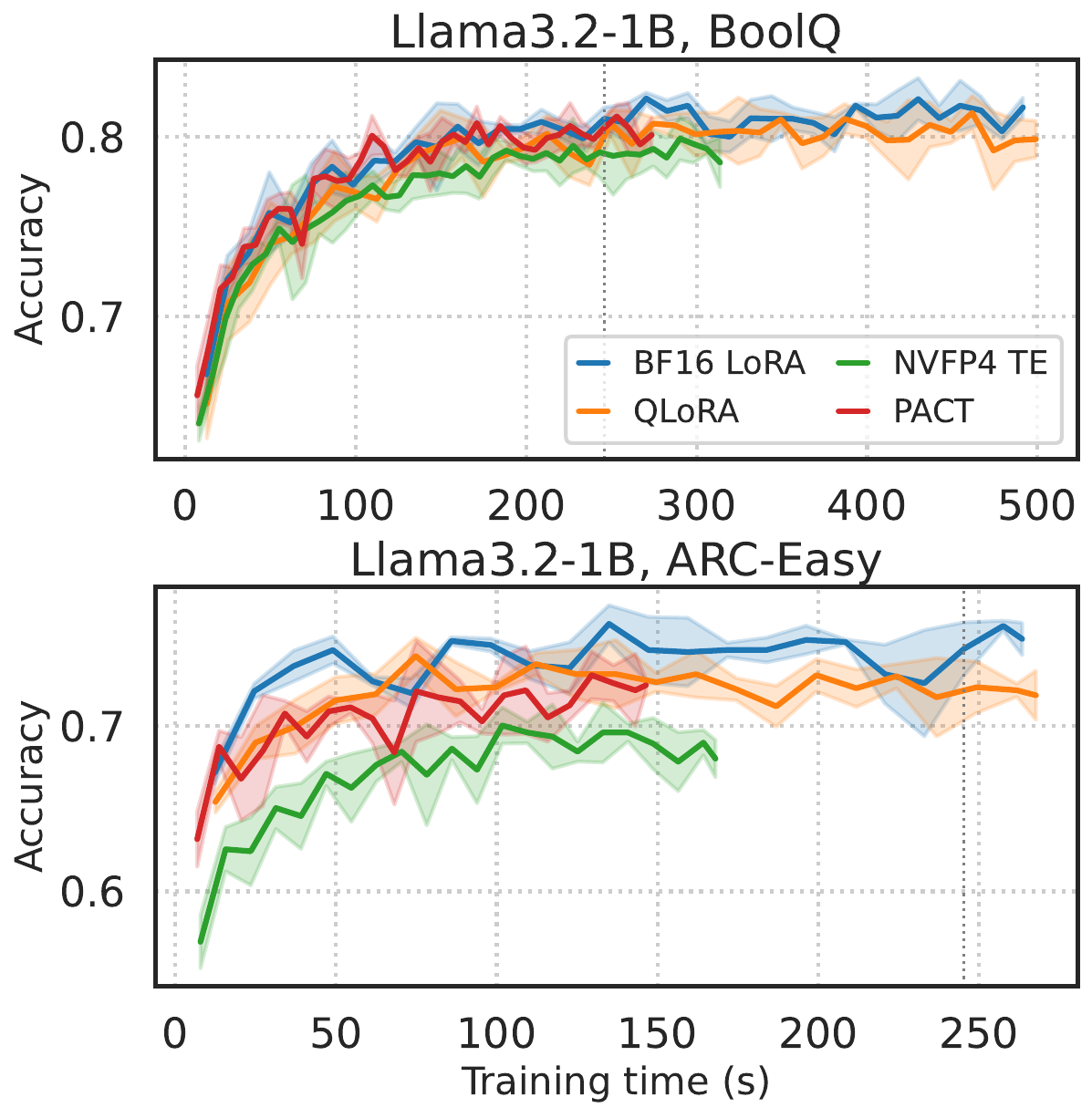}
\caption{Llama-1B accuracy versus wall-clock training time. PACT has the lowest training time for the same number of training steps.}
\label{fig:time}
\end{figure}

\textbf{End-to-end training time.}
To evaluate whether per-step speedups translate into shorter training runs, Figure~\ref{fig:time} compares training time on Llama-1B with matched update budgets of 1,000 steps for BoolQ and 537 steps for ARC-Easy. All four methods use batch size 16 and the same task-specific learning rate, with three seeds per method. Sequences are padded to the longest sequence in each batch, so the padded sequence length may vary across training steps. Time is measured as the number of updates multiplied by the median step time, excluding evaluation and offline calibration. PACT has the lowest wall-clock training time on both tasks, indicating that its operator-level acceleration can shorten training runs.

\section{Related Work}
\label{sec:related}

\textbf{Mixed-Precision Training.}
The seminal work of \citet{micikevicius2018mixed} introduced FP16-based mixed-precision training: a forward pass casts higher-precision weights and activations to lower-precision at the GEMM boundary, executes the linear-layer GEMM in lower-precision, and writes back a higher-precision output. Recent works~\citep{micikevicius2022fp8,deepseekai2024deepseekv3,nvidia2025pretrainingnvfp4} extend the approach to FP8 and FP4. Quantizing GEMM operands alone does not necessarily reduce the memory occupied by saved activations. PACT combines native NVFP4 GEMMs with NVFP4 storage of selected backward state.

\textbf{Post-Training Quantization (PTQ) and Quantization-Aware Training (QAT).}
PTQ~\citep{dettmers2022llmint8,frantar2023gptq,xiao2023smoothquant,lin2024awq} converts pretrained models to low-bit representations for efficient inference, typically using a small calibration dataset. QAT~\citep{jacob2018integer,nagel2021whitepaper,liu2024llmqat,chen2025efficientqat} instead simulates quantization during training so that parameters adapt to rounding error. Fake-quantization QAT retains higher-precision GEMMs and adds quantization overhead, rather than directly accelerating training. Both primarily target deployment efficiency. PACT uses offline calibration to prepare frozen weights, then executes base-projection GEMMs directly in NVFP4 during fine-tuning.

\textbf{Memory-Efficient Edge Training.}
\citet{lin2022device} restrict trainable parameters to selected layers and channels, enabling on-device training within 256\,KB. PockEngine~\citep{zhu2023pockengine} similarly reduces backward computation through sparse updates. FwdLLM~\citep{xu2023fwdllm} avoids storing backward activations through backpropagation-free gradient estimation, while MobiZO~\citep{gao2025mobizo} uses parallelized zeroth-order optimization. These approaches reduce training work or change gradient estimation. PACT retains first-order LoRA optimization and targets computation and storage precision. HyC-LoRA~\citep{hyclora2025} reduces retained activation memory through hybrid compression, but the computation is still performed in higher-precision.

\section{Conclusion}
PACT accelerates LoRA fine-tuning with native NVFP4 GEMMs and reduces memory usage by compressing activations retained for backward. Its data flow is designed to reduce quantization and dequantization overhead while limiting quantization error across forward and backward computation. Experiments on a 16\,GB RTX~5080 demonstrate faster training and lower memory usage with competitive adaptation quality across language and vision-language tasks. These results support practical model adaptation on memory-constrained consumer GPUs.

\clearpage
\section*{Limitations}
This work focuses on SLMs in the $1$B to $3$B range, where activation memory is a major bottleneck for LoRA fine-tuning on memory-constrained edge hardware. Scaling PACT to larger models may introduce additional numerical and systems challenges. PACT currently retains attention computation in BF16. Extending NVFP4 computation to FlashAttention is a direction for future work. Our evaluation also uses a single memory-constrained GPU. Multi-GPU and datacenter settings, where communication, sharding, and pipeline parallelism alter the trade-offs, remain unexplored. Finally, PACT targets transformer decoders with LoRA adapters. Extending its low-precision computation and activation storage to other architectures and adaptation methods remains future work. This work develops an efficiency method for SLM fine-tuning and does not introduce new model capabilities or deployment risks.

\bibliography{custom}

@inproceedings{hu2022lora,
  title     = {{LoRA}: Low-Rank Adaptation of Large Language Models},
  author    = {Hu, Edward J. and Shen, Yelong and Wallis, Phillip and Allen-Zhu, Zeyuan and Li, Yuanzhi and Wang, Shean and Wang, Lu and Chen, Weizhu},
  booktitle = {International Conference on Learning Representations (ICLR)},
  year      = {2022},
  url = {https://openreview.net/forum?id=nZeVKeeFYf9}
}

@inproceedings{dettmers2023qlora,
  title     = {{QLoRA}: Efficient Finetuning of Quantized {LLMs}},
  author    = {Dettmers, Tim and Pagnoni, Artidoro and Holtzman, Ari and Zettlemoyer, Luke},
  booktitle = {Advances in Neural Information Processing Systems (NeurIPS)},
  year      = {2023},
  url = {https://proceedings.neurips.cc/paper_files/paper/2023/hash/1feb87871436031bdc0f2beaa62a049b-Abstract.html},
  volume = {36},
  doi = {10.52202/075280-0441},
  pages = {10088--10115}
}

@article{deepseekai2024deepseekv3,
  title   = {{DeepSeek-V3} Technical Report},
  author  = {{DeepSeek-AI}},
  journal = {arXiv preprint arXiv:2412.19437},
  year    = {2024},
  url = {https://arxiv.org/abs/2412.19437}
}

@article{nvidia2025pretrainingnvfp4,
  title   = {Pretraining Large Language Models with {NVFP4}},
  author  = {{NVIDIA}},
  journal = {arXiv preprint arXiv:2509.25149},
  year    = {2025},
  url = {https://arxiv.org/abs/2509.25149}
}

@misc{nvidiaBlackwellArchitecture,
  title        = {{NVIDIA RTX Blackwell GPU Architecture}},
  author       = {{NVIDIA}},
  url          = {https://images.nvidia.com/aem-dam/Solutions/geforce/blackwell/nvidia-rtx-blackwell-gpu-architecture.pdf},
  year         = {2025}
}

@article{rouhani2023microscaling,
  title   = {Microscaling Data Formats for Deep Learning},
  author  = {Rouhani, Bita Darvish and Zhao, Ritchie and More, Ankit and Hall, Mathew and Khodamoradi, Alireza and Deng, Summer and Choudhary, Dhruv and Cornea, Marius and Dellinger, Eric and Denolf, Kristof and Stosic, Dusan and Elango, Venmugil and Golub, Maximilian and Heinecke, Alexander and James-Roxby, Phil and Jani, Dharmesh and Kolhe, Gaurav and Langhammer, Martin and Li, Ada and Melnick, Levi and Mesmakhosroshahi, Maral and Rodriguez, Andres and Schulte, Michael and Shafipour, Rasoul and Shao, Lei and Siu, Michael and Dubey, Pradeep and Micikevicius, Paulius and Naumov, Maxim and Verrilli, Colin and Wittig, Ralph and Burger, Doug and Chung, Eric},
  journal = {arXiv preprint arXiv:2310.10537},
  year    = {2023},
  url = {https://arxiv.org/abs/2310.10537}
}

@inproceedings{micikevicius2018mixed,
  title     = {Mixed Precision Training},
  author    = {Micikevicius, Paulius and Narang, Sharan and Alben, Jonah and Diamos, Gregory and Elsen, Erich and Garcia, David and Ginsburg, Boris and Houston, Michael and Kuchaiev, Oleksii and Venkatesh, Ganesh and Wu, Hao},
  booktitle = {International Conference on Learning Representations (ICLR)},
  year      = {2018},
  url = {https://openreview.net/forum?id=r1gs9JgRZ}
}

@article{micikevicius2022fp8,
  title   = {{FP8} Formats for Deep Learning},
  author  = {Micikevicius, Paulius and Stosic, Dusan and Burgess, Neil and Cornea, Marius and Dubey, Pradeep and Grisenthwaite, Richard and Ha, Sangwon and Heinecke, Alexander and Judd, Patrick and Kamalu, John and Mellempudi, Naveen and Oberman, Stuart and Shoeybi, Mohammad and Siu, Michael and Wu, Hao},
  journal = {arXiv preprint arXiv:2209.05433},
  year    = {2022},
  url = {https://arxiv.org/abs/2209.05433}
}

@inproceedings{dettmers2022llmint8,
  title     = {{LLM.int8()}: 8-bit Matrix Multiplication for Transformers at Scale},
  author    = {Dettmers, Tim and Lewis, Mike and Belkada, Younes and Zettlemoyer, Luke},
  booktitle = {Advances in Neural Information Processing Systems (NeurIPS)},
  year      = {2022},
  url = {https://proceedings.neurips.cc/paper_files/paper/2022/hash/c3ba4962c05c49636d4c6206a97e9c8a-Abstract-Conference.html},
  volume = {35},
  doi = {10.52202/068431-2198},
  pages = {30318--30332}
}

@inproceedings{frantar2023gptq,
  title     = {{GPTQ}: Accurate Post-Training Quantization for Generative Pre-trained Transformers},
  author    = {Frantar, Elias and Ashkboos, Saleh and Hoefler, Torsten and Alistarh, Dan},
  booktitle = {International Conference on Learning Representations (ICLR)},
  year      = {2023},
  url = {https://openreview.net/forum?id=tcbBPnfwxS}
}

@inproceedings{xiao2023smoothquant,
  title     = {{SmoothQuant}: Accurate and Efficient Post-Training Quantization for Large Language Models},
  author    = {Xiao, Guangxuan and Lin, Ji and Seznec, Mickael and Wu, Hao and Demouth, Julien and Han, Song},
  booktitle = {International Conference on Machine Learning},
  year      = {2023},
  url = {https://proceedings.mlr.press/v202/xiao23c.html},
  pages = {38087--38099}
}

@inproceedings{lin2024awq,
  title     = {{AWQ}: Activation-aware Weight Quantization for On-Device {LLM} Compression and Acceleration},
  author    = {Lin, Ji and Tang, Jiaming and Tang, Haotian and Yang, Shang and Chen, Wei-Ming and Wang, Wei-Chen and Xiao, Guangxuan and Dang, Xingyu and Gan, Chuang and Han, Song},
  booktitle = {Proceedings of Machine Learning and Systems (MLSys)},
  year      = {2024},
  url = {https://proceedings.mlsys.org/paper_files/paper/2024/hash/42a452cbafa9dd64e9ba4aa95cc1ef21-Abstract-Conference.html},
  volume = {6},
  pages = {87--100}
}

@inproceedings{jacob2018integer,
  title     = {Quantization and Training of Neural Networks for Efficient Integer-Arithmetic-Only Inference},
  author    = {Jacob, Benoit and Kligys, Skirmantas and Chen, Bo and Zhu, Menglong and Tang, Matthew and Howard, Andrew and Adam, Hartwig and Kalenichenko, Dmitry},
  booktitle = {Proceedings of the IEEE Conference on Computer Vision and Pattern Recognition},
  year      = {2018},
  url = {https://openaccess.thecvf.com/content_cvpr_2018/html/Jacob_Quantization_and_Training_CVPR_2018_paper.html},
  pages = {2704--2713}
}

@article{nagel2021whitepaper,
  title   = {A White Paper on Neural Network Quantization},
  author  = {Nagel, Markus and Fournarakis, Marios and Amjad, Rana Ali and Bondarenko, Yelysei and van Baalen, Mart and Blankevoort, Tijmen},
  journal = {arXiv preprint arXiv:2106.08295},
  year    = {2021},
  url = {https://arxiv.org/abs/2106.08295}
}

@inproceedings{liu2024llmqat,
  title     = {{LLM-QAT}: Data-Free Quantization Aware Training for Large Language Models},
  author    = {Liu, Zechun and Oguz, Barlas and Zhao, Changsheng and Chang, Ernie and Stock, Pierre and Mehdad, Yashar and Shi, Yangyang and Krishnamoorthi, Raghuraman and Chandra, Vikas},
  booktitle = {Findings of the Association for Computational Linguistics: ACL 2024},
  year      = {2024},
  url = {https://aclanthology.org/2024.findings-acl.26/},
  doi = {10.18653/v1/2024.findings-acl.26},
  pages = {467--484}
}

@inproceedings{chen2025efficientqat,
  title     = {{EfficientQAT}: Efficient Quantization-Aware Training for Large Language Models},
  author    = {Chen, Mengzhao and Shao, Wenqi and Xu, Peng and Wang, Jiahao and Gao, Peng and Zhang, Kaipeng and Luo, Ping},
  booktitle = {Proceedings of the 63rd Annual Meeting of the Association for Computational Linguistics (Volume 1: Long Papers)},
  year      = {2025},
  url = {https://aclanthology.org/2025.acl-long.498/},
  doi = {10.18653/v1/2025.acl-long.498},
  pages = {10081--10100}
}

@inproceedings{lin2022device,
  title     = {On-Device Training Under 256{KB} Memory},
  author    = {Lin, Ji and Zhu, Ligeng and Chen, Wei-Ming and Wang, Wei-Chen and Gan, Chuang and Han, Song},
  booktitle = {Advances in Neural Information Processing Systems (NeurIPS)},
  year      = {2022},
  url = {https://proceedings.neurips.cc/paper_files/paper/2022/hash/90c56c77c6df45fc8e556a096b7a2b2e-Abstract-Conference.html},
  volume = {35},
  doi = {10.52202/068431-1667},
  pages = {22941--22954}
}

@inproceedings{zhu2023pockengine,
  title     = {{PockEngine}: Sparse and Efficient Fine-tuning in a Pocket},
  author    = {Zhu, Ligeng and Hu, Lanxiang and Lin, Ji and Wang, Wei-Chen and Chen, Wei-Ming and Gan, Chuang and Han, Song},
  booktitle = {Proceedings of the 56th Annual IEEE/ACM International Symposium on Microarchitecture (MICRO)},
  year      = {2023},
  doi = {10.1145/3613424.3614307},
  url = {https://doi.org/10.1145/3613424.3614307}
}

@article{xu2023fwdllm,
  title   = {{FwdLLM}: Efficient {FedLLM} Using Forward Gradient},
  author  = {Xu, Mengwei and Cai, Dongqi and Wu, Yaozong and Li, Xiang and Wang, Shangguang},
  journal = {arXiv preprint arXiv:2308.13894},
  year    = {2023},
  url = {https://arxiv.org/abs/2308.13894}
}

@inproceedings{gao2025mobizo,
  title     = {{MobiZO}: Enabling Efficient {LLM} Fine-Tuning at the Edge via Inference Engines},
  author    = {Gao, Lei and Ziashahabi, Amir and Niu, Yue and Avestimehr, Salman and Annavaram, Murali},
  booktitle = {Proceedings of the 2025 Conference on Empirical Methods in Natural Language Processing},
  year      = {2025},
  url = {https://aclanthology.org/2025.emnlp-main.1022/},
  doi = {10.18653/v1/2025.emnlp-main.1022},
  pages = {20206--20223}
}

@article{grattafiori2024llama3,
  title   = {The {Llama 3} Herd of Models},
  author  = {Grattafiori, Aaron and Dubey, Abhimanyu and Jauhri, Abhinav and Pandey, Abhinav and Kadian, Abhishek and Al-Dahle, Ahmad and Letman, Aiesha and Mathur, Akhil and Schelten, Alan and Vaughan, Alex and Yang, Amy and Fan, Angela and Goyal, Anirudh and Hartshorn, Anthony and Yang, Aobo and Mitra, Archi and Sravankumar, Archie and Korenev, Artem and Hinsvark, Arthur and Rao, Arun and Zhang, Aston and Rodriguez, Aurelien and Gregerson, Austen and Spataru, Ava and Roziere, Baptiste and Biron, Bethany and Tang, Binh and Chern, Bobbie and Caucheteux, Charlotte and Nayak, Chaya and Bi, Chloe and Marra, Chris and McConnell, Chris and Keller, Christian and Touret, Christophe and Wu, Chunyang and Wong, Corinne and Ferrer, Cristian Canton and Nikolaidis, Cyrus and Allonsius, Damien and Song, Daniel and Pintz, Danielle and Livshits, Danny and Wyatt, Danny and Esiobu, David and Choudhary, Dhruv and Mahajan, Dhruv and Garcia-Olano, Diego and Perino, Diego and Hupkes, Dieuwke and Lakomkin, Egor and AlBadawy, Ehab and Lobanova, Elina and Dinan, Emily and Smith, Eric Michael and Radenovic, Filip and Guzm{\'a}n, Francisco and Zhang, Frank and Synnaeve, Gabriel and Lee, Gabrielle and Anderson, Georgia Lewis and Thattai, Govind and Nail, Graeme and Mialon, Gregoire and Pang, Guan and Cucurell, Guillem and Nguyen, Hailey and Korevaar, Hannah and Xu, Hu and Touvron, Hugo and Zarov, Iliyan and Ibarra, Imanol Arrieta and Kloumann, Isabel and Misra, Ishan and Evtimov, Ivan and Zhang, Jack and Copet, Jade and Lee, Jaewon and Geffert, Jan and Vranes, Jana and Park, Jason and Mahadeokar, Jay and Shah, Jeet and van der Linde, Jelmer and Billock, Jennifer and Hong, Jenny and Lee, Jenya and Fu, Jeremy and Chi, Jianfeng and Huang, Jianyu and Liu, Jiawen and Wang, Jie and Yu, Jiecao and Bitton, Joanna and Spisak, Joe and Park, Jongsoo and Rocca, Joseph and Johnstun, Joshua and Saxe, Joshua and Jia, Junteng and Alwala, Kalyan Vasuden and Prasad, Karthik and Upasani, Kartikeya and Plawiak, Kate and Li, Ke and Heafield, Kenneth and Stone, Kevin and El-Arini, Khalid and Iyer, Krithika and Malik, Kshitiz and Chiu, Kuenley and Bhalla, Kunal and Lakhotia, Kushal and Rantala-Yeary, Lauren and van der Maaten, Laurens and Chen, Lawrence and Tan, Liang and Jenkins, Liz and Martin, Louis and Madaan, Lovish and Malo, Lubo and Blecher, Lukas and Landzaat, Lukas and de Oliveira, Luke and Muzzi, Madeline and Pasupuleti, Mahesh and Singh, Mannat and Paluri, Manohar and Kardas, Marcin and Tsimpoukelli, Maria and Oldham, Mathew and Rita, Mathieu and Pavlova, Maya and Kambadur, Melanie and Lewis, Mike and Si, Min and Singh, Mitesh Kumar and Hassan, Mona and Goyal, Naman and Torabi, Narjes and Bashlykov, Nikolay and Bogoychev, Nikolay and Chatterji, Niladri and Zhang, Ning and Duchenne, Olivier and {\c{C}}elebi, Onur and Alrassy, Patrick and Zhang, Pengchuan and Li, Pengwei and Vasic, Petar and Weng, Peter and Bhargava, Prajjwal and Dubal, Pratik and Krishnan, Praveen and Koura, Punit Singh and Xu, Puxin and He, Qing and Dong, Qingxiao and Srinivasan, Ragavan and Ganapathy, Raj and Calderer, Ramon and Cabral, Ricardo Silveira and Stojnic, Robert and Raileanu, Roberta and Maheswari, Rohan and Girdhar, Rohit and Patel, Rohit and Sauvestre, Romain and Polidoro, Ronnie and Sumbaly, Roshan and Taylor, Ross and Silva, Ruan and Hou, Rui and Wang, Rui and Hosseini, Saghar and Chennabasappa, Sahana and Singh, Sanjay and Bell, Sean and Kim, Seohyun Sonia and Edunov, Sergey and Nie, Shaoliang and Narang, Sharan and Raparthy, Sharath and Shen, Sheng and Wan, Shengye and Bhosale, Shruti and Zhang, Shun and Vandenhende, Simon and Batra, Soumya and Whitman, Spencer and Sootla, Sten and Collot, Stephane and Gururangan, Suchin and Borodinsky, Sydney and Herman, Tamar and Fowler, Tara and Sheasha, Tarek and Georgiou, Thomas and Scialom, Thomas and Speckbacher, Tobias and Mihaylov, Todor and Xiao, Tong and Karn, Ujjwal and Goswami, Vedanuj and Gupta, Vibhor and Ramanathan, Vignesh and Kerkez, Viktor and Gonguet, Vincent and Do, Virginie and Vogeti, Vish and Albiero, V{\'i}tor and Petrovic, Vladan and Chu, Weiwei and Xiong, Wenhan and Fu, Wenyin and Meers, Whitney and Martinet, Xavier and Wang, Xiaodong and Wang, Xiaofang and Tan, Xiaoqing Ellen and Xia, Xide and Xie, Xinfeng and Jia, Xuchao and Wang, Xuewei and Goldschlag, Yaelle and Gaur, Yashesh and Babaei, Yasmine and Wen, Yi and Song, Yiwen and Zhang, Yuchen and Li, Yue and Mao, Yuning and Coudert, Zacharie Delpierre and Yan, Zheng and Chen, Zhengxing and Papakipos, Zoe and Singh, Aaditya and Srivastava, Aayushi and Jain, Abha and Kelsey, Adam and Shajnfeld, Adam and Gangidi, Adithya and Victoria, Adolfo and Goldstand, Ahuva and Menon, Ajay and Sharma, Ajay and Boesenberg, Alex and Baevski, Alexei and Feinstein, Allie and Kallet, Amanda and Sangani, Amit and Teo, Amos and Yunus, Anam and Lupu, Andrei and Alvarado, Andres and Caples, Andrew and Gu, Andrew and Ho, Andrew and Poulton, Andrew and Ryan, Andrew and Ramchandani, Ankit and Dong, Annie and Franco, Annie and Goyal, Anuj and Saraf, Aparajita and Chowdhury, Arkabandhu and Gabriel, Ashley and Bharambe, Ashwin and Eisenman, Assaf and Yazdan, Azadeh and James, Beau and Maurer, Ben and Leonhardi, Benjamin and Huang, Bernie and Loyd, Beth and De Paola, Beto and Paranjape, Bhargavi and Liu, Bing and Wu, Bo and Ni, Boyu and Hancock, Braden and Wasti, Bram and Spence, Brandon and Stojkovic, Brani and Gamido, Brian and Montalvo, Britt and Parker, Carl and Burton, Carly and Mejia, Catalina and Liu, Ce and Wang, Changhan and Kim, Changkyu and Zhou, Chao and Hu, Chester and Chu, Ching-Hsiang and Cai, Chris and Tindal, Chris and Feichtenhofer, Christoph and Gao, Cynthia and Civin, Damon and Beaty, Dana and Kreymer, Daniel and Li, Daniel and Adkins, David and Xu, David and Testuggine, Davide and David, Delia and Parikh, Devi and Liskovich, Diana and Foss, Didem and Wang, Dingkang and Le, Duc and Holland, Dustin and Dowling, Edward and Jamil, Eissa and Montgomery, Elaine and Presani, Eleonora and Hahn, Emily and Wood, Emily and Le, Eric-Tuan and Brinkman, Erik and Arcaute, Esteban and Dunbar, Evan and Smothers, Evan and Sun, Fei and Kreuk, Felix and Tian, Feng and Kokkinos, Filippos and Ozgenel, Firat and Caggioni, Francesco and Kanayet, Frank and Seide, Frank and Florez, Gabriela Medina and Schwarz, Gabriella and Badeer, Gada and Swee, Georgia and Halpern, Gil and Herman, Grant and Sizov, Grigory and Guangyi and Zhang and Lakshminarayanan, Guna and Inan, Hakan and Shojanazeri, Hamid and Zou, Han and Wang, Hannah and Zha, Hanwen and Habeeb, Haroun and Rudolph, Harrison and Suk, Helen and Aspegren, Henry and Goldman, Hunter and Zhan, Hongyuan and Damlaj, Ibrahim and Molybog, Igor and Tufanov, Igor and Leontiadis, Ilias and Veliche, Irina-Elena and Gat, Itai and Weissman, Jake and Geboski, James and Kohli, James and Lam, Janice and Asher, Japhet and Gaya, Jean-Baptiste and Marcus, Jeff and Tang, Jeff and Chan, Jennifer and Zhen, Jenny and Reizenstein, Jeremy and Teboul, Jeremy and Zhong, Jessica and Jin, Jian and Yang, Jingyi and Cummings, Joe and Carvill, Jon and Shepard, Jon and McPhie, Jonathan and Torres, Jonathan and Ginsburg, Josh and Wang, Junjie and Wu, Kai and U, Kam Hou and Saxena, Karan and Khandelwal, Kartikay and Zand, Katayoun and Matosich, Kathy and Veeraraghavan, Kaushik and Michelena, Kelly and Li, Keqian and Jagadeesh, Kiran and Huang, Kun and Chawla, Kunal and Huang, Kyle and Chen, Lailin and Garg, Lakshya and A, Lavender and Silva, Leandro and Bell, Lee and Zhang, Lei and Guo, Liangpeng and Yu, Licheng and Moshkovich, Liron and Wehrstedt, Luca and Khabsa, Madian and Avalani, Manav and Bhatt, Manish and Mankus, Martynas and Hasson, Matan and Lennie, Matthew and Reso, Matthias and Groshev, Maxim and Naumov, Maxim and Lathi, Maya and Keneally, Meghan and Liu, Miao and Seltzer, Michael L. and Valko, Michal and Restrepo, Michelle and Patel, Mihir and Vyatskov, Mik and Samvelyan, Mikayel and Clark, Mike and Macey, Mike and Wang, Mike and Hermoso, Miquel Jubert and Metanat, Mo and Rastegari, Mohammad and Bansal, Munish and Santhanam, Nandhini and Parks, Natascha and White, Natasha and Bawa, Navyata and Singhal, Nayan and Egebo, Nick and Usunier, Nicolas and Mehta, Nikhil and Laptev, Nikolay Pavlovich and Dong, Ning and Cheng, Norman and Chernoguz, Oleg and Hart, Olivia and Salpekar, Omkar and Kalinli, Ozlem and Kent, Parkin and Parekh, Parth and Saab, Paul and Balaji, Pavan and Rittner, Pedro and Bontrager, Philip and Roux, Pierre and Dollar, Piotr and Zvyagina, Polina and Ratanchandani, Prashant and Yuvraj, Pritish and Liang, Qian and Alao, Rachad and Rodriguez, Rachel and Ayub, Rafi and Murthy, Raghotham and Nayani, Raghu and Mitra, Rahul and Parthasarathy, Rangaprabhu and Li, Raymond and Hogan, Rebekkah and Battey, Robin and Wang, Rocky and Howes, Russ and Rinott, Ruty and Mehta, Sachin and Siby, Sachin and Bondu, Sai Jayesh and Datta, Samyak and Chugh, Sara and Hunt, Sara and Dhillon, Sargun and Sidorov, Sasha and Pan, Satadru and Mahajan, Saurabh and Verma, Saurabh and Yamamoto, Seiji and Ramaswamy, Sharadh and Lindsay, Shaun and Lindsay, Shaun and Feng, Sheng and Lin, Shenghao and Zha, Shengxin Cindy and Patil, Shishir and Shankar, Shiva and Zhang, Shuqiang and Zhang, Shuqiang and Wang, Sinong and Agarwal, Sneha and Sajuyigbe, Soji and Chintala, Soumith and Max, Stephanie and Chen, Stephen and Kehoe, Steve and Satterfield, Steve and Govindaprasad, Sudarshan and Gupta, Sumit and Deng, Summer and Cho, Sungmin and Virk, Sunny and Subramanian, Suraj and Choudhury, Sy and Goldman, Sydney and Remez, Tal and Glaser, Tamar and Best, Tamara and Koehler, Thilo and Robinson, Thomas and Li, Tianhe and Zhang, Tianjun and Matthews, Tim and Chou, Timothy and Shaked, Tzook and Vontimitta, Varun and Ajayi, Victoria and Montanez, Victoria and Mohan, Vijai and Kumar, Vinay Satish and Mangla, Vishal and Ionescu, Vlad and Poenaru, Vlad and Mihailescu, Vlad Tiberiu and Ivanov, Vladimir and Li, Wei and Wang, Wenchen and Jiang, Wenwen and Bouaziz, Wes and Constable, Will and Tang, Xiaocheng and Wu, Xiaojian and Wang, Xiaolan and Wu, Xilun and Gao, Xinbo and Kleinman, Yaniv and Chen, Yanjun and Hu, Ye and Jia, Ye and Qi, Ye and Li, Yenda and Zhang, Yilin and Zhang, Ying and Adi, Yossi and Nam, Youngjin and Yu and Wang and Zhao, Yu and Hao, Yuchen and Qian, Yundi and Li, Yunlu and He, Yuzi and Rait, Zach and DeVito, Zachary and Rosnbrick, Zef and Wen, Zhaoduo and Yang, Zhenyu and Zhao, Zhiwei and Ma, Zhiyu},
  journal = {arXiv preprint arXiv:2407.21783},
  year    = {2024},
  url = {https://arxiv.org/abs/2407.21783}
}

@article{yang2024qwen25,
  title   = {{Qwen2.5} Technical Report},
  author  = {Yang, An and Yang, Baosong and Zhang, Beichen and Hui, Binyuan and Zheng, Bo and Yu, Bowen and Li, Chengyuan and Liu, Dayiheng and Huang, Fei and Wei, Haoran and Lin, Huan and Yang, Jian and Tu, Jianhong and Zhang, Jianwei and Yang, Jianxin and Yang, Jiaxi and Zhou, Jingren and Lin, Junyang and Dang, Kai and Lu, Keming and Bao, Keqin and Yang, Kexin and Yu, Le and Li, Mei and Xue, Mingfeng and Zhang, Pei and Zhu, Qin and Men, Rui and Lin, Runji and Li, Tianhao and Tang, Tianyi and Xia, Tingyu and Ren, Xingzhang and Ren, Xuancheng and Fan, Yang and Su, Yang and Zhang, Yichang and Wan, Yu and Liu, Yuqiong and Cui, Zeyu and Zhang, Zhenru and Qiu, Zihan},
  journal = {arXiv preprint arXiv:2412.15115},
  year    = {2024},
  url = {https://arxiv.org/abs/2412.15115}
}

@inproceedings{wang2019glue,
  title     = {{GLUE}: A Multi-Task Benchmark and Analysis Platform for Natural Language Understanding},
  author    = {Wang, Alex and Singh, Amanpreet and Michael, Julian and Hill, Felix and Levy, Omer and Bowman, Samuel R.},
  booktitle = {International Conference on Learning Representations (ICLR)},
  year      = {2019},
  url = {https://openreview.net/forum?id=rJ4km2R5t7}
}

@inproceedings{wang2019superglue,
  title     = {{SuperGLUE}: A Stickier Benchmark for General-Purpose Language Understanding Systems},
  author    = {Wang, Alex and Pruksachatkun, Yada and Nangia, Nikita and Singh, Amanpreet and Michael, Julian and Hill, Felix and Levy, Omer and Bowman, Samuel R.},
  booktitle = {Advances in Neural Information Processing Systems (NeurIPS)},
  year      = {2019},
  url = {https://proceedings.neurips.cc/paper/2019/hash/4496bf24afe7fab6f046bf4923da8de6-Abstract.html},
  volume = {32}
}

@article{clark2018arc,
  title   = {Think you have Solved Question Answering? Try {ARC}, the {AI2} Reasoning Challenge},
  author  = {Clark, Peter and Cowhey, Isaac and Etzioni, Oren and Khot, Tushar and Sabharwal, Ashish and Schoenick, Carissa and Tafjord, Oyvind},
  journal = {arXiv preprint arXiv:1803.05457},
  year    = {2018},
  url = {https://arxiv.org/abs/1803.05457}
}

@article{sakaguchi2021winogrande,
  title   = {{WinoGrande}: An Adversarial Winograd Schema Challenge at Scale},
  author  = {Sakaguchi, Keisuke and Le Bras, Ronan and Bhagavatula, Chandra and Choi, Yejin},
  journal = {Communications of the ACM},
  volume  = {64},
  number  = {9},
  pages   = {99--106},
  year    = {2021},
  doi = {10.1145/3474381},
  url = {https://doi.org/10.1145/3474381}
}

@misc{nvidiaTransformerEngine2026,
  title        = {{NVIDIA Transformer Engine}},
  author       = {{NVIDIA}},
  url          = {https://github.com/NVIDIA/TransformerEngine},
  year         = {2026}
}

@article{xu2024ondevice,
  title   = {On-Device Language Models: A Comprehensive Review},
  author  = {Xu, Jiajun and Li, Zhiyuan and Chen, Wei and Wang, Qun and Gao, Xin and Cai, Qi and Ling, Ziyuan},
  journal = {arXiv preprint arXiv:2409.00088},
  year    = {2024},
  url = {https://arxiv.org/abs/2409.00088}
}

@article{gholami2021survey,
  title   = {A Survey of Quantization Methods for Efficient Neural Network Inference},
  author  = {Gholami, Amir and Kim, Sehoon and Dong, Zhen and Yao, Zhewei and Mahoney, Michael W. and Keutzer, Kurt},
  journal = {arXiv preprint arXiv:2103.13630},
  year    = {2021},
  url = {https://arxiv.org/abs/2103.13630}
}

@article{bai2025qwen25vl,
  title   = {{Qwen2.5-VL} Technical Report},
  author  = {Bai, Shuai and Chen, Keqin and Liu, Xuejing and Wang, Jialin and Ge, Wenbin and Song, Sibo and Dang, Kai and Wang, Peng and Wang, Shijie and Tang, Jun and Zhong, Humen and Zhu, Yuanzhi and Yang, Mingkun and Li, Zhaohai and Wan, Jianqiang and Wang, Pengfei and Ding, Wei and Fu, Zheren and Xu, Yiheng and Ye, Jiabo and Zhang, Xi and Xie, Tianbao and Cheng, Zesen and Zhang, Hang and Yang, Zhibo and Xu, Haiyang and Lin, Junyang},
  journal = {arXiv preprint arXiv:2502.13923},
  year    = {2025},
  url = {https://arxiv.org/abs/2502.13923}
}

@inproceedings{lu2022scienceqa,
  title     = {Learn to Explain: Multimodal Reasoning via Thought Chains for Science Question Answering},
  author    = {Lu, Pan and Mishra, Swaroop and Xia, Tony and Qiu, Liang and Chang, Kai-Wei and Zhu, Song-Chun and Tafjord, Oyvind and Clark, Peter and Kalyan, Ashwin},
  booktitle = {Advances in Neural Information Processing Systems (NeurIPS)},
  year      = {2022},
  url = {https://proceedings.neurips.cc/paper_files/paper/2022/hash/11332b6b6cf4485b84afadb1352d3a9a-Abstract-Conference.html},
  volume = {35},
  doi = {10.52202/068431-0182},
  pages = {2507--2521}
}

@inproceedings{hyclora2025,
  title     = {{HyC-LoRA}: Memory Efficient {LoRA} Fine-tuning with Hybrid Activation Compression},
  author    = {Wang, Yujin and Dong, Shunan and Huang, Zongle and You, Yichen and He, Liu and Yang, Huazhong and Liu, Yongpan and Jia, Hongyang},
  booktitle = {Proceedings of Machine Learning and Systems (MLSys)},
  year      = {2025},
  url = {https://proceedings.mlsys.org/paper_files/paper/2025/hash/5431dca75a8d2abc1fb51e89e8324f10-Abstract-Conference.html},
  volume = {7}
}

@inproceedings{rajpurkar2016squad,
  title     = {{SQuAD}: 100,000+ Questions for Machine Comprehension of Text},
  author    = {Rajpurkar, Pranav and Zhang, Jian and Lopyrev, Konstantin and Liang, Percy},
  booktitle = {Proceedings of the 2016 Conference on Empirical Methods in Natural Language Processing},
  year      = {2016},
  url = {https://aclanthology.org/D16-1264/},
  doi = {10.18653/v1/D16-1264},
  pages = {2383--2392}
}

@article{cobbe2021gsm8k,
 title = {Training Verifiers to Solve Math Word Problems},
 author = {Cobbe, Karl and Kosaraju, Vineet and Bavarian, Mohammad and Chen, Mark and Jun, Heewoo and Kaiser, Lukasz and Plappert, Matthias and Tworek, Jerry and Hilton, Jacob and Nakano, Reiichiro and Hesse, Christopher and Schulman, John},
 journal = {arXiv preprint arXiv:2110.14168},
 year = {2021},
 url = {https://arxiv.org/abs/2110.14168}
}

@inproceedings{gliwa2019samsum,
 title = {{SAMSum} Corpus: A Human-annotated Dialogue Dataset for Abstractive Summarization},
 author = {Gliwa, Bogdan and Mochol, Iwona and Biesek, Maciej and Wawer, Aleksander},
 booktitle = {Proceedings of the 2nd Workshop on New Frontiers in Summarization},
 year = {2019},
 url = {https://aclanthology.org/D19-5409/},
  doi = {10.18653/v1/D19-5409},
  pages = {70--79}
}

@inproceedings{tan2024oppu,
  title     = {Democratizing Large Language Models via Personalized Parameter-Efficient Fine-tuning},
  author    = {Tan, Zhaoxuan and Zeng, Qingkai and Tian, Yijun and Liu, Zheyuan and Yin, Bing and Jiang, Meng},
  booktitle = {Proceedings of the 2024 Conference on Empirical Methods in Natural Language Processing},
  year      = {2024},
  month     = nov,
  pages     = {6476--6491},
  doi       = {10.18653/v1/2024.emnlp-main.372},
  url       = {https://aclanthology.org/2024.emnlp-main.372/}
}
\clearpage
\appendix

\section{Comparison of MXFP4 and NVFP4}
\label{app:fp4_compare}

MXFP4 and NVFP4 pair four-bit E2M1 values with a shared block scale. The E2M1 codebook contains 15 distinct values,
\begin{equation}
\{0,\,\pm0.5,\,\pm1,\,\pm1.5,\,\pm2,\,\pm3,\,\pm4,\,\pm6\}.
\end{equation}
Table~\ref{tab:fp4_compare} compares their block sizes and scale formats. Including the eight-bit block scale, MXFP4 requires $4+8/32=4.25$ bits per element, and NVFP4 requires $4+8/16=4.5$ bits per element. This accounting excludes padding and the additional FP32 tensor scale used for weights and backward gradients in PACT.

\begin{table}[h]
\centering\small
\begin{tabular}{lcc}
\toprule
Property & MXFP4 & NVFP4 \\
\midrule
Block size & 32 & 16 \\
Element format & E2M1 & E2M1 \\
Block-scale format & E8M0 & E4M3 \\
Bits per element & 4.25 & 4.50 \\
\bottomrule
\end{tabular}
\caption{MXFP4 and NVFP4 format parameters, showing that NVFP4 uses smaller groups and more precise block scales at a slightly higher storage cost.}
\label{tab:fp4_compare}
\end{table}

\textbf{Block size.}
Sharing a scale among fewer elements limits the number of values affected by an outlier within a block. PACT uses NVFP4's groups of 16 along the reduction dimension of each GEMM, allowing adjacent groups to use different scales.

\textbf{Scale precision.}
E8M0 represents power-of-two scales, while E4M3 provides three mantissa bits within each exponent. NVFP4 therefore permits finer scale adjustment, which PACT accounts for when selecting weight block scales. NVFP4 activation storage is approximately $4.5/16=28\%$ of BF16 before padding and auxiliary statistics.

\section{Implementation Details}
\label{app:implementation}

\subsection{Precision and Saved State}
\label{app:precision}

Table~\ref{tab:precision} separates the formats used for forward computation, retained state, and backward computation. All seven frozen projections in every decoder layer use NVFP4 operands. Nonlinear arithmetic and reductions use FP32, while attention, residuals, and adapter computation retain BF16 tensors. This precision allocation applies to the Q/V-only adapter configuration used in our main experiments.

\begin{table*}[t]
\centering\small
\begin{tabularx}{\textwidth}{lXXX}
\toprule
Component & Forward computation & State retained for backward & Backward computation \\
\midrule
Frozen projections & NVFP4 GEMMs with FP32 accumulation and BF16 output & Cached forward and backward NVFP4 weight layouts & NVFP4 input-gradient GEMMs \\
RMSNorm & BF16 input and FP32 normalization & NVFP4 output and FP32 per-token normalization statistic & FP32 reconstruction and reduction, BF16 gradient \\
SiLU-multiply & BF16 gate/up inputs and FP32 arithmetic & Separate NVFP4 gate/up copies & FP32 reconstruction and derivatives, BF16 gradients \\
Attention and RoPE & BF16 tensors & Standard attention backward state & Standard attention backward path \\
LoRA branch & BF16 GEMMs on reconstructed inputs & Shared NVFP4 input and BF16 adapter parameters & BF16 adapter gradients and input gradients \\
Residuals & BF16 addition & As required by the surrounding operators & BF16 gradient accumulation \\
Embeddings and LM head & Higher-precision path & As required by the original operators & Original operators and fused cross-entropy \\
\bottomrule
\end{tabularx}
\caption{PACT's precision allocation, showing how native NVFP4 projections and compressed backward state coexist with higher-precision computation.}
\label{tab:precision}
\end{table*}

The quantized pre-attention RMSNorm output is shared by the Q/K/V projections, the Q/V adapters, and RMSNorm backward. The MLP retains its quantized RMSNorm output and gate/up copies for the nonlinear backward kernels. The down-projection input can be released after forward because the frozen down projection needs only its weight and output gradient during backward. The quantized input to the frozen attention output projection can also be released after forward, although the BF16 attention output remains part of the attention backward state. These lifetimes explain the retained-state terms in Equation~\ref{eq:memory}.

\subsection{Quantizer and Weight Preparation}
\label{app:quantizer}

\textbf{Weight scales.}
For a frozen weight $\mathbf{W}_0$, PACT uses the FP32 tensor scale $s_W=\max|\mathbf{W}_0|/(448\cdot6)$, with a positive floor for zero-valued inputs. Each row is partitioned into groups of 16 along the forward reduction dimension. The scale search in Section~\ref{sec:weights} evaluates nine factors: 1.5, 1.25, 1.0, 0.95, 0.90, 0.85, 0.80, 0.75, and 0.70. Each candidate scale is rounded to UE4M3 before measuring reconstruction error. The selected group scale remains fixed during subsequent column quantization and error compensation.

\textbf{Calibration settings.}
GPTQ uses processing blocks of 128 columns, groups of 16 columns, and diagonal damping equal to 1\% of the mean Hessian diagonal. Projections are processed in the order Q/K/V, attention output, gate/up, and down. Text calibration uses 128 sequences of length 512 sampled from the WikiText-2 training split with seed zero. The input Hessian is accumulated from reconstructed NVFP4 activations, so weight calibration uses the activation precision of the forward GEMMs. ScienceQA image-prompt calibration uses 128 training prompts containing images. Weight preparation is performed once and cached for subsequent fine-tuning runs.

\textbf{Backward layout.}
PACT reconstructs the calibrated forward weight, transposes it, casts it to BF16, and applies round-to-nearest NVFP4 quantization along the new reduction dimension. The resulting representation $\tilde{\mathbf{W}}_{0,b}$ has its own tensor and block scales. Both packed layouts are cached and reused throughout fine-tuning. Their reconstructed values may differ because their quantization groups follow different dimensions, so the base input-gradient GEMM approximates the backward pass of the forward weight representation.

\textbf{Activation and gradient scales.}
The fused RMSNorm and SiLU-multiply forward kernels use block scales without an additional tensor scale in the default configuration. Before an input-gradient GEMM, the backward quantizer uses
\begin{equation}
s_{\nabla Y}=\frac{\max|\nabla\mathbf{Y}|}{448\cdot6}
\end{equation}
with a positive floor, followed by UE4M3 scales for groups of 16 values. The GEMM epilogue applies the product of the gradient and weight tensor scales. Keeping these scales on the device avoids a host-side scalar read during each GEMM. Tensor scaling adapts the quantizer to the gradient range, although sufficiently small values can still round to zero.

\subsection{Nonlinear Backward Computation}
\label{app:backward}

For RMSNorm, the saved FP32 statistic is
\begin{equation}
r_x=\left(\frac{1}{d}\sum_{i=1}^{d}x_i^2+\epsilon\right)^{-1/2}.
\end{equation}
With frozen normalization weights $\boldsymbol{\gamma}$, PACT reconstructs the normalized input from the saved output as $\hat{\mathbf{x}}\simeq\bar{\mathbf{y}}/\boldsymbol{\gamma}$ and evaluates Equation~\ref{eq:rms_backward}. This removes the separate retained input tensor. For a zero normalization weight, the implementation sets the corresponding reconstructed coordinate to zero to avoid division by zero. The reconstruction is approximate because the retained output is quantized.

For SiLU-multiply, let $\sigma$ denote the sigmoid function. The backward kernel reconstructs $\bar{\mathbf{g}}$ and $\bar{\mathbf{u}}$ from their saved NVFP4 copies and evaluates
\begin{align}
\nabla\mathbf{g}&\simeq\nabla\mathbf{z}\odot\bar{\mathbf{u}}\odot\sigma(\bar{\mathbf{g}})
\odot[1+\bar{\mathbf{g}}\odot(1-\sigma(\bar{\mathbf{g}}))],\\
\nabla\mathbf{u}&\simeq\nabla\mathbf{z}\odot\bar{\mathbf{g}}\odot\sigma(\bar{\mathbf{g}}).
\end{align}
Reconstruction and derivative evaluation take place within the same kernel, which writes BF16 gradients. Subsequent input-gradient GEMMs quantize these gradients using the scaling scheme in Appendix~\ref{app:quantizer}.

\subsection{Operator Algorithms}
\label{app:algorithms}

Algorithms~\ref{alg:pact_nonlinear} and~\ref{alg:pact_lora_linear} summarize the nonlinear and LoRA linear operators. Quantized tensors include their scales. A nonlinear operator consumes BF16 inputs, performs arithmetic in FP32, and writes packed NVFP4 outputs. Its retained state is chosen for the backward expressions in Appendix~\ref{app:backward}.

\begin{algorithm*}[t]
\caption{PACT nonlinear forward operations produce NVFP4 projection inputs while retaining compressed backward state.}
\label{alg:pact_nonlinear}
\begin{algorithmic}[1]
\Procedure{RMSNormForward}{$\mathbf{X},\boldsymbol{\gamma},\epsilon$}
    \State Read BF16 $\mathbf{X}$ and compute each token's inverse RMS $r_x$ in FP32
    \State Compute $\mathbf{Y}\gets\boldsymbol{\gamma}\odot(r_x\mathbf{X})$ in FP32
    \State Quantize and store $\tilde{\mathbf{Y}}\gets\Q(\mathbf{Y})$ for the subsequent projections
    \State Retain $\tilde{\mathbf{Y}}$ and $r_x$ for backward
    \State \Return $\tilde{\mathbf{Y}}$
\EndProcedure
\Statex
\Procedure{SiLUMultiplyForward}{$\mathbf{g},\mathbf{u}$}
    \State Read BF16 gate/up outputs $\mathbf{g},\mathbf{u}$
    \State Compute $\mathbf{z}\gets\operatorname{SiLU}(\mathbf{g})\odot\mathbf{u}$ in FP32
    \State Quantize and store $\tilde{\mathbf{z}}\gets\Q(\mathbf{z})$ for the down projection
    \State Quantize and retain $\tilde{\mathbf{g}}\gets\Q(\mathbf{g})$ and $\tilde{\mathbf{u}}\gets\Q(\mathbf{u})$ for backward
    \State Release the BF16 gate/up outputs after their forward use
    \State \Return $\tilde{\mathbf{z}}$
\EndProcedure
\end{algorithmic}
\end{algorithm*}

\begin{algorithm*}[t]
\caption{PACT LoRA linear forward and backward operations combine NVFP4 base GEMMs with BF16 adapter computation.}
\label{alg:pact_lora_linear}
\begin{algorithmic}[1]
\Require NVFP4 input $\tilde{\mathbf{X}}$, cached NVFP4 weights $\tilde{\mathbf{W}}_0,\tilde{\mathbf{W}}_{0,b}$, BF16 adapters $\mathbf{A},\mathbf{B}$
\Statex \textbf{Forward}
\State $\mathbf{Y}_{\mathrm{base}}\gets\tilde{\mathbf{X}}\tilde{\mathbf{W}}_0^\top$ using an NVFP4 GEMM
\State Reconstruct $\bar{\mathbf{X}}\gets\DQ(\tilde{\mathbf{X}})$ in BF16 for the adapter branch
\State $\mathbf{Y}\gets\mathbf{Y}_{\mathrm{base}}+(\bar{\mathbf{X}}\mathbf{A}^\top)\mathbf{B}^\top$ using BF16 adapter GEMMs
\State Retain the shared NVFP4 input $\tilde{\mathbf{X}}$ for adapter gradients
\State \Return $\mathbf{Y}$
\Statex \textbf{Backward, given BF16 $\nabla\mathbf{Y}$}
\State Quantize $\widetilde{\nabla\mathbf{Y}}\gets\Q(\nabla\mathbf{Y})$ with tensor and block scales
\State $\nabla\mathbf{X}_{\mathrm{base}}\gets\widetilde{\nabla\mathbf{Y}}\tilde{\mathbf{W}}_{0,b}$ using an NVFP4 GEMM
\State Reconstruct $\bar{\mathbf{X}}\gets\DQ(\tilde{\mathbf{X}})$ in BF16
\State $\nabla\mathbf{X}\gets\nabla\mathbf{X}_{\mathrm{base}}+(\nabla\mathbf{Y}\mathbf{B})\mathbf{A}$
\State $\nabla\mathbf{B}\gets(\nabla\mathbf{Y})^\top(\bar{\mathbf{X}}\mathbf{A}^\top)$
\State $\nabla\mathbf{A}\gets(\nabla\mathbf{Y}\mathbf{B})^\top\bar{\mathbf{X}}$
\State \Return $\nabla\mathbf{X},\nabla\mathbf{A},\nabla\mathbf{B}$
\end{algorithmic}
\end{algorithm*}

The implementation pads token rows to multiples of 128 and stores scales in the layout required by the NVFP4 GEMM kernels. RMSNorm first computes the per-token statistic, then normalizes and quantizes its output. SiLU-multiply produces its quantized output and two retained input copies in one fused kernel. Nonlinear outputs are quantized before they are written to global memory, avoiding a separate higher-precision output buffer.

The adapter GEMMs contract with the rank-$r$ factors first, as shown in Algorithm~\ref{alg:pact_lora_linear}. This keeps intermediate tensors low-rank and gives adapter computation a cost proportional to $bsr(d_{\text{in}}+d_{\text{out}})$ without materializing a dense adapter weight update. Frozen projections without adapters omit the adapter computations and do not retain their input solely for weight-gradient computation.

\section{Training and Evaluation Protocol}
\label{app:protocol}

\textbf{Compute budget.}
We estimate the total computational cost of this study at approximately 200 GPU hours on an RTX~5080, including calibration, hyperparameter search, training, evaluation, and profiling.

\subsection{Task Formulation and Prompts}
\label{app:prompts}

We formulate fine-tuning as conditional generation and compute next-token cross-entropy only on the target tokens. Table~\ref{tab:template} summarizes the input and target structures. Classification uses short verbalized labels, while COPA and WinoGrande use candidate continuations. Generation tasks use an answer, reasoning solution, or summary as the target. Prompt and padding positions are masked out of the loss.

\begin{table*}[t]
\centering\small
\begin{tabularx}{\textwidth}{lXl}
\toprule
Task & Input structure & Target \\
\midrule
SST-2 & Sentence followed by ``It was'' & terrible / great \\
RTE & Premise and a question asking whether the hypothesis is true & Yes / No \\
BoolQ & Passage followed by the question & Yes / No \\
WSC & Text and a question about the pronoun's referent & Yes / No \\
WiC & Word, two sentences, and a question about shared meaning & Yes / No \\
COPA & Premise followed by ``because'' or ``so'' & Candidate continuation \\
WinoGrande & Sentence prefix before the blank & Option and sentence suffix \\
ARC-Easy / ARC-Challenge & Question, labeled options, and ``Answer:'' & Answer letter \\
SQuAD & Title, context, question, and ``Answer:'' & Short answer \\
GSM8K & Question followed by ``Answer:'' & Solution and final number \\
SAMSum & Dialogue followed by ``Summary:'' & Dialogue summary \\
ScienceQA & Image, question, and labeled options & Answer letter \\
\bottomrule
\end{tabularx}
\caption{Task input and target structures, showing how classification, generation, and vision-language adaptation share a target-token training objective.}
\label{tab:template}
\end{table*}

\subsection{Classification and Multiple Choice}
\label{app:classification}

The main classification protocol uses rank-16 adapters on Q/V projections with adapter scaling equal to one. Fine-tuning uses AdamW and a constant learning rate selected from the sweep in Section~\ref{sec:setup}. Only adapter parameters are updated. Dataset order and subsampling follow seeds 42, 43, and 44. We first average the scores for a learning rate across seeds, then select the best mean for each task and method. The Avg. column of Table~\ref{tab:accuracy} weights the nine tasks equally, and zero-shot scores are reported before adaptation.
For label-based tasks, evaluation selects the most likely candidate answer token. COPA and WinoGrande compare the mean log-probability of target tokens for each candidate continuation, with paired candidates kept together. 

\subsection{Open-Ended Generation}
\label{app:generation}

Generation runs use 500 optimizer steps, up to 4,000 training examples, and 500 evaluation examples per seed. We select the learning rate with the highest mean score among rates with completed runs for all three seeds, using candidates $10^{-4}$ and $5\times10^{-4}$. Evaluation is greedy, with no sampling and one beam. SQuAD generates at most 50 tokens and is scored with normalized token F1. GSM8K generates at most 320 tokens and uses exact match of the extracted final number. SAMSum generates at most 80 tokens and uses ROUGE-L F1. Generation stops at the configured newline or end-of-sequence token.


\subsection{Vision-Language Adaptation}
\label{app:vlm_protocol}

ScienceQA uses image-bearing training and validation examples and evaluates answer-letter accuracy on 1,000 examples. The image encoder and merger remain frozen, and adaptation changes only the decoder adapters. PACT uses micro-batch size 4 with four accumulation steps, while BF16 uses micro-batch size 2 with eight accumulation steps, giving the same effective batch size of 16. Runs train for 500 steps. Text calibration is evaluated with learning rates $5\times10^{-5}$, $10^{-4}$, and $5\times10^{-4}$. The image-prompt calibration result averages three runs at $5\times10^{-4}$. Calibration prompts come from the training split.

\section{Comparison with Activation-Memory Reduction Methods}
\label{app:memory_alternatives}

We compare PACT with decoder-layer checkpointing and HyC-LoRA~\citep{hyclora2025} to evaluate the memory savings and execution costs of recomputation and activation compression.

\subsection{Gradient Checkpointing}
\label{app:checkpoint}

Table~\ref{tab:checkpoint} compares Llama-1B configurations at batch size 4 and sequence length 1,024. Checkpointed BF16 LoRA uses less memory than PACT alone but takes substantially longer per step. Combining PACT with checkpointing reduces peak memory to 2.27\,GB, using 28\% less memory and running $1.95\times$ faster than checkpointed BF16 LoRA. Activation compression and recomputation therefore provide complementary benefits.

\begin{table}[t]
\centering\small
\setlength{\tabcolsep}{4pt}
\begin{tabular}{lrr}
\toprule
Llama-1B & Step (ms) & Peak (GB) \\
\midrule
BF16 LoRA & 240 & 7.06 \\
BF16 + checkpointing & 310 & 3.14 \\
PACT & 125 & 3.72 \\
PACT + checkpointing & 159 & 2.27 \\
\bottomrule
\end{tabular}
\caption{Training time and peak memory with decoder-layer checkpointing.}
\label{tab:checkpoint}
\end{table}

Checkpointing uses non-reentrant recomputation for each decoder layer. At sequence length 1,024, it increases the largest tested fitting batch from 10 to 78 for BF16 LoRA and from 26 to 88 for PACT. These capacity gains do not imply proportional throughput improvements.

\subsection{HyC-LoRA}
\label{app:hyclora}

We evaluate the released HyC-LoRA implementation on Llama-1B using the same RTX~5080, comparing each method with its own uncompressed reference at batch size 4 and sequence length 512. Two-bit HyC-LoRA reduces allocated memory by 37\% but increases step time by 13\% (Table~\ref{tab:hyclora}). PACT reduces peak memory by 44\% and step time by 45\% relative to BF16 LoRA. These within-implementation results show that PACT combines memory savings with faster execution by reusing quantized activations and accelerating frozen projections with native NVFP4 GEMMs. HyC-LoRA also enables larger workloads, including $8\!\times\!512$ and $4\!\times\!1024$, that exceed its uncompressed reference's memory capacity. Absolute values are not directly comparable across the two methods because their implementations and configurations differ.

\begin{table}[t]
\centering\small
\setlength{\tabcolsep}{3pt}
\begin{tabular}{llrr}
\toprule
Shape & HyC-LoRA setting & MiB & ms \\
\midrule
$4\!\times\!512$ & Own baseline & 9,706 & 194 \\
& 2-bit compression & 6,112 & 219 \\
& 4-bit compression & 6,341 & n/a \\
\midrule
$8\!\times\!512$ & Own baseline & OOM & n/a \\
& 2-bit compression & 9,627 & 425 \\
& 4-bit compression & 10,088 & n/a \\
\midrule
$4\!\times\!1024$ & Own baseline & OOM & n/a \\
& 2-bit compression & 10,140 & 490 \\
& 4-bit compression & 10,607 & n/a \\
\bottomrule
\end{tabular}
\caption{HyC-LoRA allocated memory and step time. Shapes denote batch size $\times$ sequence length. The abbreviation n/a indicates unmeasured timing.}
\label{tab:hyclora}
\end{table}

\section{Training-Step Time Breakdown}
\label{app:profile}

Table~\ref{tab:step_profile} profiles Llama-1B at batch size 4 and sequence length 1,024. PACT shifts frozen-projection computation to NVFP4, substantially reducing GEMM time. Attention, higher-precision operations, and quantization overhead remain, limiting the whole-step speedup relative to individual NVFP4 GEMMs.

\begin{table}[h]
\centering\small
\begin{tabular}{lrr}
\toprule
CUDA-time category (ms) & BF16 LoRA & PACT \\
\midrule
BF16 GEMMs & 202.7 & 48.9 \\
NVFP4 GEMMs & 0 & 25.6 \\
Quantization / fused kernels & 0 & 17.1 \\
Attention & 12.9 & 12.7 \\
Elementwise / other & 21.5 & 21.8 \\
\midrule
Reported profile total & 239.4 & 130.1 \\
\bottomrule
\end{tabular}
\caption{Training-step time breakdown for Llama-1B.}
\label{tab:step_profile}
\end{table}

The BF16 GEMM category includes decoder projections, adapters, and the output head. Profile totals differ from the uninstrumented measurements in Figure~\ref{fig:fixed}, and category times need not sum exactly to the reported total.

\end{document}